%% file: ms.tex
\documentclass[twoside,11pt]{article}

\usepackage[abbrv, preprint]{jmlr2e}
\usepackage{times,enumitem,float,amsmath,amsfonts,amssymb}
\usepackage{hyperref}
\usepackage{mathtools}
\usepackage{pifont}
\setlist[enumerate]{wide=0pt, widest=99,leftmargin=15pt, labelsep=*}
\newtheorem{problem}[theorem]{Problem} %[section]

\usepackage{color,subcaption}
\usepackage{booktabs}

\usepackage{tikz}
\usepackage{wrapfig}
\usepackage{pstricks}
\usepackage{pgf}
\usepackage{pdfpages, fp}
\usepackage{epic,bez123}
\usepackage{floatflt}

\usepackage{arydshln}

\usepackage{ifpdf}

\usepackage{wrapfig}% package for wrapfigure environment

\jmlrheading{1}{2021}{}{2/21}{}{hrvt21}{Christian Horvat and Jean-Pascal Pfister}

\ShortHeadings{Density estimation on low-dimensional manifolds: an inflation-deflation approach}{Horvat and Pfister}
\firstpageno{1}

\begin{document}

\title{Density estimation on low-dimensional manifolds: an inflation-deflation approach}

\author{\name Christian Horvat \email christian.horvat@unibe.ch \\
       \addr Department of Physiology\\
       University of Bern\\
       Bern, Switzerland
       \AND
       \name Jean-Pascal Pfister \email jeanpascal.pfister@unibe.ch \\
       \addr Department of Physiology\\
       University of Bern\\
       Bern, Switzerland}

\editor{ }

\maketitle

\begin{abstract}%   <- trailing '%' for backward compatibility of .sty file
Normalizing Flows (NFs) are universal density estimators based on Neural Networks. However, this universality is limited: the density's support needs to be diffeomorphic to a Euclidean space. In this paper, we propose a novel method to overcome this limitation without sacrificing universality. The proposed method inflates the data manifold by adding noise in the normal space, trains an NF on this inflated manifold, and, finally, deflates the learned density. Our main result provides sufficient conditions on the manifold and the specific choice of noise under which the corresponding estimator is exact. Our method has the same computational complexity as NFs and does not require computing an inverse flow. We also show that, if the embedding dimension is much larger than the manifold dimension, noise in the normal space can be well approximated by Gaussian noise. This allows using our method for approximating arbitrary densities on unknown manifolds provided that the manifold dimension is known. 
\end{abstract}

\begin{keywords}
Normalizing Flow, Density Estimation, low-dimensional manifolds, normal space, noise
\end{keywords}

\input{JMLR_introduction}

%\input{JMLR_problem}
\input{JMLR_problem_new}

\input{JMLR_methods}
\input{JMLR_related_work}
\input{JMLR_results}
\input{JMLR_discussion}
\newpage
\acks{We would like to thank Johann Brehmer for clarifying discussions on the manifold flow, and Simone C. Surace for useful discussions on manifolds. \\
	This study has been supported by the Swiss National Science Foundation grant 31003A\_175644.}

\appendix

%\bibliography{iclr2021_conference}

% Acknowledgements should go at the end, before appendices and references


%\newpage

\input{JMLR_appendix}
\input{JMLR_appendix2}
\newpage

\bibliography{ms}

\end{document}

%% file: JMLR_introduction.tex
\section{Introduction}\label{introduction}
 Many modern problems involving high-dimensional data are formulated probabilistically. Key concepts, such as Bayesian Classification, Denoising, or Anomaly Detection, rely on the data generating density $p^*(x)$. Therefore, a main research area and of crucial importance is learning this data generating density  $p^*(x)$ from samples. 
 
 For the case where the corresponding random variable $X$ with values in $\mathbb{R}^D$ takes values on a manifold diffeomorphic to $\mathbb{R}^D$, a Normalizing Flow (NF)  can be used to learn $p^*(x)$ exactly (\citet{huang2018neural}). However, in practice, many real-world applications such as predicting protein structures in molecular biology (\cite{hamelryck2006sampling}), learning  motions in robotics (\cite{feiten2013rigid}), or predicting earthquake patterns in geology (\cite{geller1997earthquake}) are modeled on low-dimensional manifolds, and therefore gave rise to the manifold hypothesis which states that high-dimensional datasets, such as high-resolution images, live close to a  low-dimensional manifold (see \cite{fefferman2016testing} and the references therein). As a consequence, few attempts have been made to use NFs to learn densities on low-dimensional manifolds, overcoming their topological constraint. To do so, these methods either need to know the manifold beforehand (\citet{gemici2016normalizing}, \citet{rezende2020normalizing}, \citet{mathieu2020riemannian}, \citet{lou2020neural}), or sacrifice the directness of the estimate (\citet{beitler2018pie}, \citet{kim2020softflow}, \citet{cunningham2020normalizing}, \citet{brehmer2020flows}). 

 Our goal in this paper is to overcome both the aforementioned limitations of using NFs for density estimation on Riemannian manifolds. Given data points from a $d-$dimensional Riemannian manifold denoted as $\mathcal{X}$ embedded in $\mathbb{R}^D$, $d<D$, we first inflate the manifold by adding a specific noise in the normal space direction of the manifold, then train an NF on this inflated manifold, and, finally, deflate the trained density by exploiting the choice of  noise and the geometry of the manifold. See Figure \ref{fig:intuition} for a schematic overview of these points. 
 \input{intuition_icls.TpX} 

 Our main theorem states sufficient conditions on the manifold and the type of noise we use for the inflation step such that the deflation becomes exact. To guarantee the exactness, we do need to know the manifold as in e.g. \citet{rezende2020normalizing} because we need to be able to sample in the manifold's normal space. However, as we will show, for the special case where $D\gg d$, the usual Gaussian noise approximates a Gaussian restricted to the normal space. This allows using our method for approximating arbitrary densities on Riemannian manifolds provided that the manifold dimension is known. In addition, our method is based on a single NF without the necessity to invert it. Hence, we don't add any additional complexity to the training procedure of NFs such that autoregressive flows (which are typically $D$-times slower to invert) can be used. To the best of our knowledge, this is the first theoretical study that provides sufficient conditions for the learnability of a density with support on a low-dimensional manifold using NFs.

\textbf{Notations:} 
%We denote the determinant of the Gram matrix of $f:\mathbb{R}^d\to \mathbb{R}^D,d<D,$ evaluated at $f^{-1}(x)$ as
%\begin{equation}\label{eq:}
%\det G_f(x):=\det \left( J_{f}(f^{-1}(x))^TJ_{f}(f^{-1}(x)) \right)
%\end{equation}
%where $J_f(f^{-1}(x))^T$ is the transposed Jacobian of $f$. 
We denote the Lebesgue measure in $\mathbb{R}^n$ as $\lambda_{n}$. Random variables will be denoted with a capital letter, e.g. $X$, and their corresponding state spaces with the calligraphic version, $\mathcal{X}$.  Small letters correspond to vectors with dimensionality given by context. The letters $d,D,n$, and $N$ are always natural numbers. 
%If not otherwise stated, the norm $||\cdot||$ represents the Euclidean one.and the identity matrix as $I_n$. 

%% file: JMLR_problem_new.tex
\section{Background and problem statement}\label{sec:problem}
%\subsection{Normalizing Flows}\label{sec:NF}
Let $X$ be a random variable that takes values on a $d-$dimensional manifold $\mathcal{X}$ embedded in $\mathbb{R}^D$,  i.e $\mathcal{X} \subset \mathbb{R}^D$, and let $X$ be generated by an unobserved random variable $U\in \mathcal{U} \subset \mathbb{R}^d$ with density $\pi_u(u)$, where $d<D$.  That is, from a generative perspective, a sample $x$ from the random variable $X$ is obtained in the following way:
	\begin{enumerate}[label=\arabic*.]
		\item sampling from the prior: $u\sim \pi_u(u)$,  
		\item mapping to the manifold: $x=f(u)$.
	\end{enumerate}
If $f:\mathcal{U}\to \mathcal{X}$ is an embedding \footnote{Thus, a regular continuously differentiable mapping (called immersion) which is, restricted to its image, a homeomorphism.} (as it is the case in \citet{gemici2016normalizing}) the density $p^*(x)$ of $X$ is given by the change of variable formula
\begin{equation}\label{eq:problem_1}
p^*(x)=\left|\det G_f(x)\right|^{-\frac12}  \pi_u(f^{-1}(x)),
\end{equation} 
where we denote the Gram matrix of $f$ evaluated at $f^{-1}(x)$ as 
\begin{equation}\label{eq:gram_det}
G_f(x):=  J_{f}(f^{-1}(x))^TJ_{f}(f^{-1}(x))  
\end{equation}
with $J_f^T$ denoting the transpose of the Jacobian of $f$.
% where $dz$ denotes the volume form associated with 
%with  and 
%\begin{align}\label{eq:}
%dV_f(x) =& \sqrt{\left| \det G_f(x) \right|} du \notag \\
%=& \sqrt{\left| \det \left( J_{f}(u)^TJ_{f}(u) \right) \ \right|} du
%\end{align}
%where $u=f^{-1}(x)$.
 Hence, given an explicit chart $f$ and samples from $p^*(x)$, we can learn the unknown density $\pi_u(u)$ using a standard NF in $\mathbb{R}^d$. However, in general, the generating function $f$ is either unknown or not an embedding creating numerical instabilities for training inputs close to singularity points.

In \citet{brehmer2020flows}, $f$ and the unknown density $\pi_u$ are learned simultaneously. Their main idea is to define $f$ as a level set of a usual flow in $\mathbb{R}^D$ and train it together with the flow in $\mathbb{R}^d$ used to learn $\pi_u$. To evaluate the density, one needs to calculate $\left|\det G_f(x)\right|^{-\frac12}$ which computational complexity is $\mathcal{O}(d^2D)+\mathcal{O}(d^3)$. Thus this approach may be slow for high-dimensional data (which we will confirm in Section \ref{sec:DE_MNIST}). Besides, to guarantee that $f$ learns the manifold they proposed several ad hoc training strategies. We tie in with the idea to use an NF for learning $p^*(x)$ with unknown $f$ and study the following problem.
\begin{problem}\label{problem}
	Let $\mathcal{X}$ be a $d-$dimensional manifold embedded in $\mathbb{R}^D$. Let $X$ be a random variable with values in $\mathcal{X}$. Given $N$ samples from $p^*(x)$ as described above, find an estimator $\hat{p}$ of $p^*$ such that in the limit of infinitely many samples we have that $\hat{p}(x) = p^*(x)$, $\mathbb{P}_X$-almost surely.
	%		\begin{equation}\label{eq:kL_pi}
	%		generated by an embedding $f:\mathbb{R}^d \to \mathbb{R}^D$ and a random variable $Z\sim \pi(z)$ in $\mathbb{R}^d$
	%		\end{equation} 
\end{problem} 
\textbf{The universality of standard NFs: }Formally, a standard NF is a  diffeomorphism $F_{\theta}:\mathcal{Z}\subseteq \mathbb{R}^D  \to \mathcal{X}\subseteq \mathbb{R}^D$ and induces a density on $\mathcal{X}$ through $p_{\theta}(x) = \left| \det G_{F_{\theta}}(x))\right|^{-\frac12} p_\mathcal{Z}(F_{\theta}^{-1}(x))$ where $p_\mathcal{Z}$ is a known density. The parameters $\theta$ are updated such that the KL-divergence between $p^{*}(x)$ and $p_{\theta}(x)$, 
\begin{equation}\label{eq:NF_objective_fct}
D_{KL}(p^*(x)||p_{\theta}(x)) = -\mathbb{E}_{x\sim p^*(x)}[\log p_{\theta}(x)] + const.
\end{equation} %p_{\theta}(x)
is minimized. For certain flow architectures, $F_{\theta}$ is expressive enough such that in the limit of infinitely hidden layers $n$, every $p^*(x)$ with support on $\mathbb{R}^D$ can be learned exactly, see  \citet{huang2018neural,huang2020augmented} for a rigorous mathematical description. However, this universality depends on the architecture and is not true for all flow types, see \citet{pmlr-v119-zhang20h}.

\begin{remark}\label{rem:densities_on_manifolds} \
	\begin{enumerate}[label=(\roman*)]
		\item Note that $p^*(x)$ is uniquely determined by the pair $(\pi,f)$. For another embedding $f'=f\circ \phi$ with $\phi$ being a diffeomorphism , the pair $(\pi',f')$ with $\pi' = \pi \circ \phi^{-1}$ induces the same density $p^*(x)$. Hence, $p^*(x)$ does not depend on the specific embedding. 
%		In general, if a manifold is embedded in $\mathbb{R}^D$, it has a natural Riemannian structure based on the Euclidean metric. This metric is invariant under diffeomorphism. More precisely, 
%		The volume form $dV_{f_1}$ induced by an embedding $f_1$ is invariant under diffeomorphisms $\phi : \mathbb{R}^d \to \mathbb{R}^d$, i.e. $\sqrt{|\det G_{f_2}(u)|}=|\det \phi(u)|\cdot \sqrt{|\det G_{f_1}(\phi(u))}$ where $f_2 = f_1\circ \phi$. Therefore, $\int_A dV_{f_1} = \int_A dV_{f_2}$ for any measurable $A\subset \mathcal{X}$ which follows immediately from the change of variable formula.
%		 Otherwise, it is not obvious which reference measure to use for densities on such manifolds (\citet{pennec2004probabilities}). This problem goes back to the Bertrand paradox for geometrical probabilities, see Chapter 1 in \citet{kendall1963geometrical}. 
%		In our case, assuming that the random variable is generated by an embedding $f_1$, the volume form $dV_{f_1}$ is a natural choice. This is because $dV_{f_1}$ is invariant under diffeomorphisms: if there is a diffeomorphism $\phi : \mathbb{R}^d \to \mathbb{R}^d$, such that $\sqrt{g^{f_2}(z)}=|\det \phi(z)|\cdot \sqrt{g^{f_1}(\phi(z))}$ where $f_2 = f_1\circ \phi$, then $\int_A dV_{f_1} = \int_A dV_{f_2}$ for any measurable $A\subset \mathcal{X}$.\footnote{This follows immediately from the change of variable formula.}
%		Hence, $p^*(x)$ is intrinsic to the manifold and does not depend on the specific embedding. 
%		In fact, $p^*(x)$ is uniquely determined by the pair $(\pi,f)$.
		\item The density $p^*(x)$ is with respect to the volume form $dV(x) = \sqrt{\left| \det G_f(x) \right|} du$, i.e. one can calculate probabilities such as $\mathbb{P}_X(A)$ for measurable $A\subset \mathcal{X}$ as follows: $\mathbb{P}(X\in A) = \int_{f^{-1}(A)} \pi_u(u) du = \int_A  p^*(x) dV(x)$. 		Viewing $p^*(x)dV$ as a differential $d-$form, we may say that the volume form  $dV$ is induced by the Euclidean metric in $\mathbb{R}^D$. 
%		dV_f(x) =& \sqrt{\left| \det G_f(x) \right|} du
%		\item An alternativ way to motivate $p^*(x)$ is via differential forms. Given the $d$-form $\omega_z = \pi(z)  dz_1 \wedge dz_2 \wedge \dots \wedge dz_d$ on $\mathbb{R}^d$ for some $z\in \mathbb{R}^d$, there is a unique $d-$form $w^*_{f(z)}$ on $M$ such that $\omega_z$ is the pullback of $\omega^*_{f(z)}$ with respect to $f$, i.e. $f^*\omega^*_{f(z)}= \omega_z$. That means, for $z_1,\dots,z_d\in \mathbb{R}^d$, we have that $\omega_z(z_1,\dots,z_d)=\omega^*_{f(z)} (J_f(z)z_1,\dots,J_f(z)z_d)$. In particular, for $z_i=e_i$, we get on the one hand $\omega_z(e_1,\dots,e_d)=\pi(z)$ and on the other hand  $\omega^*_{f(z)} (J_f(z)e_1,\dots,J_f(z)e_d) = p^*(f(z)) \sqrt{g^f(z)}$ for some function $p^*$. Thus, it must hold that $p^*(f(z))=g^f(z)^{-\frac12}\pi(z)$. From that perspective, viewing $p^*(x)dV_f$ as a differential form, $dV_f$ is the volume form induced by the euclidean metric in $\mathbb{R}^D$.
%		, which implies that $p^*(f(z))=\pi(z)(g^f(z))^{-\frac12}$.
%		
%		 with $f$ is given by $p^*dV_f = f^* \omega$. From that perspective, the volume form $dV_f$ on $\mathcal{X}$ is induced by the euclidean metric on $\mathbb{R}^D$.
%		\item \ The concept of a density on a manifold can be generalized for orientable manifolds consisting of multiple charts. Also in this case, the volume form induced by the Euclidean metric is a natural choice. %However, there is no unique generative scheme anymore.
	\end{enumerate}
\end{remark}

%% file: JMLR_methods.tex
\section{Methods}\label{methods}
To solve Problem \ref{problem}, we want to exploit the universality of NFs. We want to inflate $\mathcal{X}$ such that the inflated manifold $\tilde{\mathcal{X}}$ becomes diffeomorphic to a set $\mathcal{U}$ on which a simple density exists. By doing so, this allows us to learn the inflated density $q(\tilde{x}), \tilde{x} \in \mathbb{R}^D$, exactly using a single NF, see Section \ref{sec:problem}. Then, given such an estimator for the modified density, we approximate $p^*(x)$ and give sufficient conditions when this approximation is exact.
%approximate the original target density and give sufficient conditions when this approximation is exact.
%
%In the following, we will describe a generic way of inflating $\mathcal{X}$ and under which conditions it becomes diffeomorphic to a set $\mathcal{U}$ on which a simple density exists. This is what we call the inflation step. Doing so, allows us to learn the inflated density exactly using a single NF, see Section \ref{sec:NF}. Then, given such an estimator for the modified density, we will approximate the original target density and give sufficient conditions when this approximation is exact.
\subsection{The Inflation step }\label{ch:inflation}
Given a sample $x$ of $X$, if we add some noise $\mathcal{E} \in \mathbb{R}^D$ to it, the resulting new random variable $\tilde{X}=X+\mathcal{E}$ has the following density
\begin{equation}\label{eq:q_normal}
q(\tilde{x})=\int_{\mathcal{X}} q(\tilde{x}|x) d\mathbb{P}_X(x),
%q_{\rm{n}}(\tilde{x})=\int_{\mathcal{X}} q_{\rm{n}}(\tilde{x}|x) d\mathbb{P}_X(x) 
\end{equation}
whee $q(\tilde{x}|x)$ is the noise density. Denote the tangent space in $x$ as $T_x$ and the normal space as $N_x$. By definition, $N_x$ is the
orthogonal complement of $T_x$. Therefore, we can decompose the noise $\mathcal{E}$ into its tangent and normal component, $\mathcal{E} = \mathcal{E}_{\rm{t}}+\mathcal{E}_{\rm{n}}$. In the following, we consider noise in the normal space only, i.e. $ \mathcal{E}_{\rm{t}} = 0$, and denote the density of the resulting random variable as $q_{\rm{n}}(\tilde{x})$. The corresponding noise density $q_{\rm{n}}(\tilde{x}|x)$ has mean $x$ and domain $N_{x}$. 
%$\tilde{X}=X+\varepsilon_{\rm{n}}$ has the following density
% Denote the tangent space in $x\in \mathcal{X}$ as $T_x$ and the normal space as $N_x$.
%This tangent space is $d$-dimensional and one possible basis is given by the column vectors of the Jacbobian $J_{z}(x)$ where $z$ is a chart for a local neighbourhood in $\mathcal{X}$ of $x$. 
%By definition, $N_x$ is the orthogonal complement of $T_x$. Therefore, g
%Given a sample $x$ of $X$ with probability distribution $\mathbb{P}_X$, if we add some noise $\varepsilon \in \mathbb{R}^D$ to it, $\tilde{x}=x+\varepsilon$, we can decompose this noise into its tangent and normal component, $\varepsilon = \varepsilon_{\rm{t}} + \varepsilon_{\rm{n}}.$ 
%%We call noise in the normal space only, i.e. $\varepsilon_{\rm{t}}=0$, "normal noise" in the following. 
%The resulting new random variable $\tilde{x} = x + \varepsilon_{\rm{n}}$  has the following density
%\begin{equation}\label{eq:q_normal}
%q_{\rm{n}}(\tilde{x})=\int_{\mathcal{X}} q_{\rm{n}}(\tilde{x}|x) d\mathbb{P}_X(x) 
%\end{equation}
%where $q_{\rm{n}}(\tilde{x}|x)$ has mean $x$ and support on $N_{x}$.
 We denote the support of $q_{\rm{n}}(\cdot|x)$ by $N_{q_{\rm{n}}(\cdot|x)}$. The random variable $\tilde{X} = X +\mathcal{E}_{\rm{n}}$ is now defined on $\widetilde{\mathcal{X}} = \bigcup_{x\in \mathcal{X}}N_{q_{\rm{n}}(\cdot|x)}.$
%As sketched in the introduction of this chapter, w
We want $\widetilde{\mathcal{X}}$ to be diffeomorphic to a set $\mathcal{U}$ on which a known density can be defined. 

{From a generative perspective, a sample $\tilde{x}$ from the random variable $\widetilde{\mathcal{X}}$ is obtained in the following way:
	\begin{enumerate}[label=\arabic*.]
		\item sampling from the prior: $u\sim \pi_u(u)$ and $v \sim \pi_v(v)$
		\item mapping to the inflated manifold: $\tilde{x}_{\rm{n}} = x + A_uv$,
	\end{enumerate}
	where $\pi_v$ is the noise generating latent density in $\mathbb{R}^{D-d}$, and $A_u\in \mathbb{R}^{D}\times \mathbb{R}^{D-d}$ is the matrix with columns consisting of normal vectors spanning the normal space in $x=f(u)$. Without loss of generality, we can choose an orthonormal basis for $N_x$ such that $\det A_u^T A_u = 1$. }

%Because then we can learn $q_{\rm{n}}(\tilde{x})$ using a single NF mapping $\widetilde{\mathcal{X}}$ to $\mathcal{U}$. 
%The reason why we emphasized on the normal space component will become clear later. In fact, we will show that for certain types of manifolds, this normal space noise will allow us to deduce the original target density $p^*(x)$ exactly. 
%Let us discuss some examples. % first.
\begin{example}\label{ex:inflation}  \
\begin{enumerate}[label=(\alph*) ]%	\item
%	If we consider usual Gaussian noise, $\varepsilon_{\rm{t}}\neq0$,  $\tilde{X}$ is diffeomorphic to $\mathbb{R}^D$. We denote the inflated density, equation (\ref{eq:q_{\rm{n}}ormal}), for this case by $q(\tilde{x})$. Thus, for this Gaussian case, $q(\tilde{x})$ can be learned by a single NF $F^{-1}$ and $p_{\mathcal{U}}(u)= \mathcal{N}(u;0,I_D)$ as reference.
	
%	\item In principle, each $x\in \mathcal{X}$ can have its own specific noise. Then, it is enough to consider the full Gaussian case for only one $x\in \mathcal{X}$. In this case, $\tilde{\mathcal{X}}$ will be diffeomorphic to $\mathbb{R}^D$ and again one NF is sufficient to learn $q(\tilde{x})$.

\item Let $\mathcal{X}=S^1 =\{x\in \mathbb{R}^2 \ | \ ||x||_2=1 \}$ be the unit circle where $||\cdot||_2$ denotes the $L_2-$norm. For each $x\in S^1$ there exists $u\in [0,2\pi)$ such that $x=e_r(x)=(\cos(u),\sin(u))^T$. To sample a point $\tilde{x}$ in $N_x$, which is spanned by $e_r(x)$, we sample a scalar value $v$ and set $\tilde{x}=x+v e_r(x)$. With $V  \sim \mathrm{Uniform}[-1,1)$, we have that 
	 \begin{equation}\label{eq:}
	 \widetilde{\mathcal{X}} = \bigcup_{x\in \mathcal{X}} \{x + v e_r(x) | v \in [-1,1) \} = \{ x\in \mathbb{R}^2 \ | \ ||x||_2<2  \}
	 \end{equation}
	 which is the open disk with radius $2$. The open disk is diffeomorphic to $(0,1)\times (0,1)$. Thus, $q_{\rm{n}}(\tilde{x})$ can be learned by a single NF denoted as $F^{-1}$ and $p_{\mathcal{Z}}(u)= \mathrm{Uniform}\left( (0,1)\times (0,1) \right) $ as reference.
	 \item As in (a), we consider the unit circle. Now we set $V$ to be a shifted $\chi^2-$ distribution with support $[-1,\infty)$. Then,
	 \begin{equation}\label{eq:}
	 \widetilde{\mathcal{X}} =  \bigcup_{x\in \mathcal{X}} \{x +v e_r(x) | v \in [-1,\infty) \} =\mathbb{R}^2.
	 \end{equation}
	  Thus, $q_{\rm{n}}(\tilde{x})$ can be learned by a single NF denoted as $F^{-1}$ and $p_{\mathcal{Z}}(z)= \mathcal{N}(z;0,I_D)$ as reference.
\end{enumerate} 
Both cases can be analogously extended to higher dimensions.
\end{example}

\subsection{The Deflation step}
Equation (\ref{eq:q_normal}) defines the density of the random variable $\tilde{X}=X+\mathcal{E}$. However, if the noise $\mathcal{E}$ is added in the normal space such that for each realization $\tilde{x}$ there exist only one $x$, we show that
%Our main idea is to find conditions such that % equation (\ref{eq:q_normal}) becomes 
\begin{equation}\label{eq:main_idea}
q_{\rm{n}}(x)  = q_{\rm{n}}(x|x) p^*(x).  %=\int_{\mathcal{X}} q_{\rm{n}}(\tilde{x}|x) d\mathbb{P}_X(x)
\end{equation}
If the estimator $\hat{q}_{\rm{n}}(\tilde{x})$ is exact, i.e. $\hat{q}_{\rm{n}}(\tilde{x}) = q_{\rm{n}}(\tilde{x})$ for $\mathbb{P}_{\tilde{X}}-$almost all $\tilde{x}\in \widetilde{\mathcal{X}}$, we have for $\tilde{x}=x$ that $p^*(x) = \hat{q}_{\rm{n}}(x) / q_{\rm{n}}(x|x)$ and therefore $p^*(x)$ can be computed from an NF and a known scaling factor.
%
%, i.e. there exists a projection operator $\text{proj}:\tilde{X}\to \mathcal{X}$, $\tilde{x}\mapsto \text{proj}(\tilde{x})$ such that $\text{proj}(\tilde{x})=x$ for 
%$\mathbb{P}_{\tilde{X}}-$almost all $\tilde{x}$.

For equation (\ref{eq:main_idea}) to be true, we need to guarantee that almost every $\tilde{x}$ corresponds to only one $x\in \mathcal{X}$.  This is certainly the case whenever all the normal spaces have no intersections at all (think of a simple line in $\mathbb{R}^2$).  We can relax this assumption by allowing null-set intersections. Moreover, only those subsets of the normal spaces are of interest which are generated by the specific choice of noise $q_{\rm{n}}(\tilde{x}|x)$. Thus, only the support of $q_{\rm{n}}(\tilde{x}|x)$, denoted by $N_{q_{\rm{n}}(\cdot|x)}$, matters. 
%For instance, if $q_{\rm{n}}(\tilde{x}|x)$ is uniform, then this support corresponds to a finite subset of $N_x$, see Example \ref{ex:inflation}. 
The key concept for our main result is expressed in the following definition:

\begin{definition}\label{def:normally_sep} Let $\mathcal{X}$ be a $d-$dimensional manifold and $N_x$ the normal space in $x\in\mathcal{X}$. Let $q_{\rm{n}}(\cdot|x)$ be a density defined on $N_x$ and denote by $N_{q_{\rm{n}}(\cdot|x)}\subseteq N_x$. Denote the collection of all such densities as $Q:=\{q_{\rm{n}}(\cdot|x)\}_{x\in\mathcal{X}}$. For $\tilde{x}\in \widetilde{\mathcal{X}}$, we define the set of all possible generators of $\tilde{x}$ as $\mathcal{A}(\tilde{x}) =  \{ x' \in \mathcal{X} |   N_{q_{\rm{n}}(\cdot|x') } \ni \tilde{x}  \}$.  We say $\mathcal{X}$ is \textbf{$Q-$normally reachable} if for all $x\in \mathcal{X}$, it holds that $\mathbb{P}_{\tilde{X}|X=x}\left(  \tilde{x} \in N_x | \#  \mathcal{A}(\tilde{x}) >1   \right) =0$ where $\# \mathcal{A}(\tilde{x})$ is the cardinality of the set $\mathcal{A}(\tilde{x})$. In other words, every $\tilde{x} \in N_x$ is $\mathbb{P}_{\tilde{X}|X=x}$-almost surely determined by $x$.
\end{definition}

To familiarize with this concept, consider Figure \ref{fig:example_Q} and the following example:
\begin{example}\label{ex:neighbours} 
	For the circle in example \ref{ex:inflation}, we chose $\mathcal{E}_{\rm{n}}$ to be uniformly distributed on the half-open interval $[-1,1)$. The point $(0,0)^T$ is contained in $N_{q_{\rm{n}}(\cdot|x)}$ for all $x\in \mathcal{X}$ and thus \\$N_{q_{\rm{n}}(\cdot|x')} \cap N_{q_{\rm{n}}(\cdot|x)}  = \{ (0,0)^T\}$ for all $x\neq x'$, see Figure \ref{fig:example_Q} (middle). Hence, for any given $\tilde{x}\in N_x$ we have that $\mathcal{A}(\tilde{x})=\mathcal{X}$ if $\tilde{x}=(0,0)^T$ and $\mathcal{A}(\tilde{x})=x$ otherwise. Therefore, $\# \mathcal{A}(\tilde{x})=\infty$ if $\tilde{x}=(0,0)^T$ and $\# \mathcal{A}(\tilde{x})=1$ else.
	%##############################
%	\begin{equation}
%	\mathcal{A}(\tilde{x})= $\begin{cases}
%	\mathcal{X}& \mathrm{ if }\tilde{x}=(0,0)^T,\\
%	x & \mathrm{ else, }\\
%	\end{cases}$
%	 \mathrm{ and therefore }
%	\# \mathcal{A}(\tilde{x})= \begin{cases}
%	\infty& \mathrm{ if }\tilde{x}=(0,0)^T,\\
%	1 & \mathrm{ else.}
%	\end{cases}
%	\end{equation}
	%
	%	\end{equation}
	%##############################
	%  Since $\mathbb{P}_{\tilde{X}}$ is a measure absolutely continuous with respect to $\lambda_1$,we have 
	\\ Thus,  $\mathbb{P}_{\tilde{X}|X=x}\left[  \tilde{x} \in \tilde{\mathcal{X}} |\# \mathcal{A}(\tilde{x}) >1   \right] = \mathbb{P}_{\tilde{X}|X=x}\left[ \tilde{x} = (0,0)^T \right] =0$ for all $x\in \mathcal{X}$.
	What follows is that $\mathcal{X}$ is $Q-$normally reachable.
	
	If we were to choose  $\mathcal{E}_{\rm{n}}$ to be uniformly distributed on $[-1.5,1)$, see Figure \ref{fig:example_Q} (right),  the normal spaces would overlap and we would have that  $\mathbb{P}_{\tilde{X}|X=x}\left[  \tilde{x} \in \tilde{\mathcal{X}} |\# \mathcal{A}(\tilde{x}) >1   \right]>0$. In this case, $\mathcal{X}$ would not be $Q-$normally reachable.
	\begin{figure}[H] \label{fig:normal_1D_vM}
		\centering
		\includegraphics[width=\linewidth]{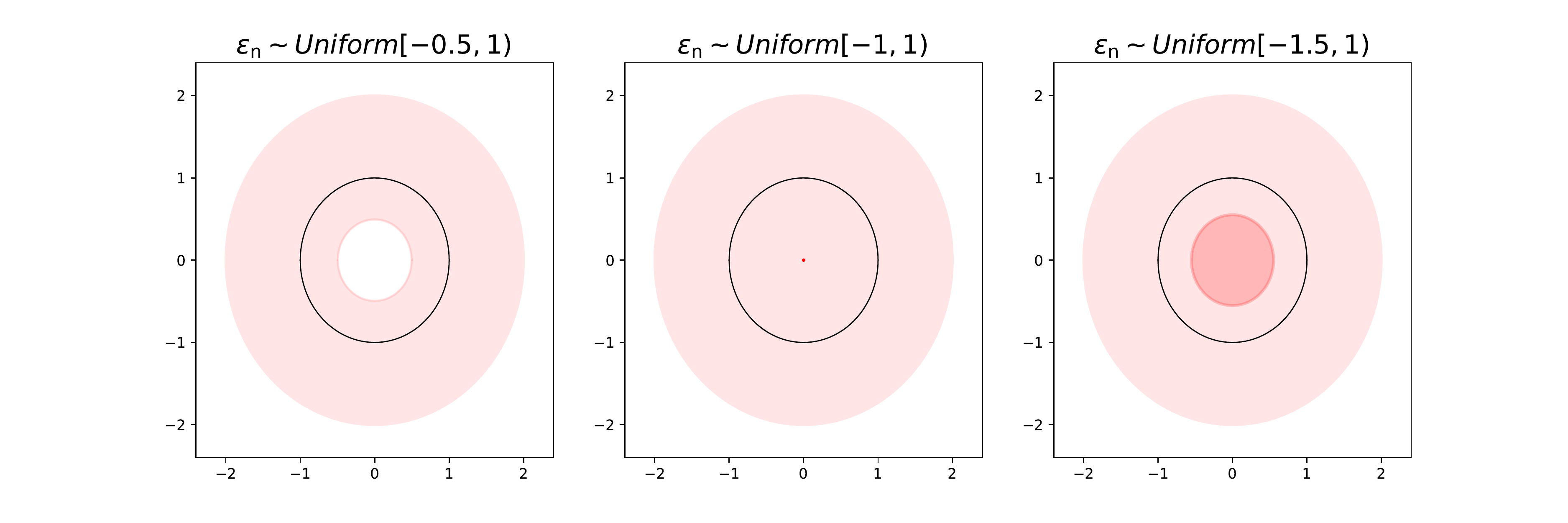}
		\caption{$Q$-normal reachability for different noise distributions $q_{\rm{n}}(\tilde{x}|x)$  used to inflate $\mathcal{X}=S^1$ (black line). \textbf{Left: }$\mathcal{X}$ is $Q$-normally reachable since every point in the inflated space $\widetilde{\mathcal{X}}$ (red shaded area) has a unique generator. \textbf{Middle: }$\mathcal{X}$ is Q-normally reachable since $\mathbb{P}_{\tilde{X}}-$almost every point in $\widetilde{\mathcal{X}}$  has a unique generator. \textbf{Right: }$\mathcal{X}$ is not Q-normally reachable since every point in the dark shaded area has two generators. Note that the pink area denotes the inflated manifold and not the density.}
		\label{fig:example_Q}
	\end{figure}
	
\end{example}

{From a generative perspective, $Q-$normal reachability ensures that the mapping
\begin{align}\label{eq:}
\tilde{f}:\mathcal{U}\times \mathcal{V} \mapsto&  \widetilde{\mathcal{X}} \notag \\
(u,v) \mapsto&f(u) + A_uv
\end{align}
is bijective (up to a set of measure $0$). As $f$ is an embedding by assumption, $\tilde{f}$ is even a diffeomorphism if $||v||_2$ is sufficiently small, as we will show in theorem \ref{thm:main}. This, together with the assumption that the prior is factorized, i.e. $\pi(u,v)=\pi_u(u)\pi_v(v)$, implies that the density $q_{\rm{n}}(\tilde{x})$ is given by %with respect to the volume form induced by $\tilde{f}$ 
\begin{equation}\label{eq:qn_factorized}
q_{\rm{n}}(\tilde{x}) = \left|\det G_{\tilde{f}}(\tilde{x})\right|^{-\frac12}  \pi_u(u) \pi_v(v)
\end{equation}
where $(u,v)=\tilde{f}^{-1}(\tilde{x})$. When setting $v=0$, we have that $\tilde{x}=x$ and indeed equation (\ref{eq:main_idea}) holds as we will show that $\left|\det G_{\tilde{f}}(x)\right|  = \left|\det G_{f} ({x})\right|$ and $\pi_v(0) = q_{\rm{n}}(x|x)$. Note that our flow of arguments does not require the manifold $\mathcal{X}$ to be generated by a single chart $f$. Hence, as long as the manifold is $Q-$normal reachable, equation (\ref{eq:qn_factorized}) holds locally for any chart $f$.
}

\begin{theorem}\label{thm:main}
	Let $\mathcal{X}$ be a $d-$dimensional, $C^2$ manifold. For each $x\in \mathcal{X}$, let $q_{\rm{n}}(\cdot|x)$ denote a continuous distribution with support $N_{q(\cdot|x)}$ in the normal space of $x$, i.e. $N_{q(\cdot|x)}\subseteq N_x$,  such that $x\in N_{q(\cdot|x)}$. Further, assume that the prior of the inflated random variable $\tilde{X}=X+\mathcal{E}_{\rm{n}}$ is factorized. If $\mathcal{X}$ is $Q-$normally reachable where $Q:=\{q_{\rm{n}}(\cdot|x)\}_{x\in\mathcal{X}}$,
%	\begin{enumerate}[label=\arabic*.]
%		\item 	,
%		\item $f\in C^2(\mathbb{R}^d)$.
%	\end{enumerate}
% and let .
%	 Let $V$ denote the volume form with respect to the product measure on $\mathcal{X}$, i.e. $V=V_f\otimes V_{h_x}$ where $f$ is the manifold generating mapping and $h_x$ the normal space generating mapping given a point $x$ on the manifold. \\	
%	Then, for $\mathbb{P}_{\tilde{X}}-$almost all $\tilde{x}\in {\mathcal{\tilde{X}}}$ holds that 
%	\begin{equation}\label{eq:}
%	\frac{d\mathbb{P}_{\tilde{X}}(\tilde{x})}{d(V_f \otimes V_{h})(\tilde{x})} = p^* (x)q_{\rm{n}}(\tilde{x}|x).
%	\end{equation}
%	Furthermore, we assume that $f\in C^2(\mathbb{R}^d)$ and $\tilde{\mathcal{X}}$ is diffeomorphic to $\mathbb{R}^D$ such that
%%	we denote the density learned using a NF (which is with respect to the volume form associated with the Euclidean metric) as $q_{\rm{n}}(\tilde{x})$ and 
%%	we assume that 
%	the density .
%	 $F$, i.e. $q_{\rm{n}}(\tilde{x})=(\det G_F(F^{-1}(\tilde{x})))^{-\frac12} p_{\mathcal{U}}(F^{-1}(\tilde{x}))$ for some known density $p_{\mathcal{U}}$. 
	then for all $x\in {\mathcal{X}}$ it holds that $q_{\rm{n}}(x) =  p^* (x)q_{\rm{n}}(x|x)$, thus 
	\begin{equation}\label{eq:main_result}
	p^* (x) =  \frac{q_{\rm{n}}(x)}{q_{\rm{n}}(x|x)}.
	\end{equation}
\end{theorem}

The proof can be found in Appendix \ref{ax:proof_thm}. As a consequence of theorem \ref{thm:main} , if the density $q_{\rm{n}}(\tilde{x})$ can be learned exactly using a single NF (which is e.g. the case whenever the inflated space $\tilde{\mathcal{X}}$ is diffeomorphic to $\mathbb{R}^D$ and the NF is sufficiently expressive), the true density $p^*(x)$ can be retrieved exactly.

\begin{proposition}\label{prop:main}	With the assumptions from theorem \ref{thm:main} , if the inflation is such that $\widetilde{\mathcal{X}}$ is diffeomorphic to $\mathbb{R}^D$, then $q_{\rm{n}}(\tilde{x})$ can be learned exactly using a single NF denoted as $F$, i.e. $q_{\rm{n}}(\tilde{x})=(\det G_F(F^{-1}(\tilde{x})))^{-\frac12} \mathcal{N}(F^{-1}(\tilde{x});0,1)$. Then, using equation (\ref{eq:main_result})  the true density $p^*(x)$ can be calculated exactly.
\end{proposition}

\begin{remark}\label{rem:product_metric}
It is important to note that the density $q_{\rm{n}}$ is with respect to the Euclidean metric because this ensures that we can learn it using an NF. If we consider the density with respect to the product metric on $\widetilde{\mathcal{X}}$ denoted as $q^{\otimes}_{\rm{n}}$, we can't use a standard NF to learn it. However, we prove in the Appendix \ref{ax:proof_product_metric} that with the assumptions of theorem \ref{thm:main} , we have that $q^{\otimes}_{\rm{n}}(\tilde{x}) = p^*(x) q_{\rm{n}}(\tilde{x}|x)$ which is based on the fact that $\widetilde{\mathcal{X}}$ isomorphic to  $\bigcup_{x\in \mathcal{X}} \left( \{x\} \times N_{q_{\rm{n}}(\cdot|x)} \right) $ up to set of measure $0$.
\end{remark}

\subsection{Gaussian noise as normal noise and the choice of $\sigma^2$}\label{sec:FG_as_NS}
Our proposed method depends on three critical points. First, we need to be able to sample in the normal space of $\mathcal{X}$. Second, we need to determine the magnitude and type of noise. Third, we need to make sure that the conditions of theorem \ref{thm:main} are fulfilled. We address (partially) those three points.
\ \\ \ \\ 
\textsl{1. }We show that for an increasing embedding dimension $D$, a full Gaussian noise is an increasingly good approximation for a Gaussian noise restricted to the normal space if we keep $d$ fixed. For that, consider $\mathcal{E} = \mathcal{E}_{\rm{t}} + \mathcal{E}_{\rm{n}}$, $\mathcal{E} \sim \mathcal{N}(0,\sigma^2 I_D)$. Then, the expected absolute squared error when approximating normal noise with full Gaussian noise is $\mathbb{E}\left[ ||\mathcal{E}-\mathcal{E}_{\rm{n}}||_2^2\right] = \mathbb{E}\left[ ||\mathcal{E}_{\rm{t}}||_2^2\right] =d\sigma^2.$ The expected relative squared error is therefore
%##############################
\begin{equation}\label{eq:normal_vs_full}
\mathbb{E}\left[ \frac{||\mathcal{E}-\mathcal{E}_{\rm{n}}||_2^2}{||\mathcal{E}_{\rm{n}}||_2^2}\right] = \mathbb{E}\left[ \frac{||\mathcal{E}_{\rm{t}}||_2^2}{||\mathcal{E}_{\rm{n}}||_2^2}\right]   =  d  \mathbb{E}\left[ \frac{\sigma^2}{||\mathcal{E}_{\rm{n}}||_2^2}\right] =  \frac{d}{D-d-2}
\end{equation}
%##############################
%\begin{align}\label{eq:}
%\mathbb{V}\left[ \frac{|\varepsilon-\varepsilon_{\rm{n}}|}{\varepsilon_{\rm{n}}}\right]  & = \mathbb{V}\left[ \frac{|\varepsilon_{\rm{t}}|}{\varepsilon_{\rm{n}}}\right]  \notag \\
%& =  d\sigma^2  \mathbb{V}\left[ \frac{1}{\varepsilon_{\rm{n}}}\right] \notag\\
%& =  \frac{d}{D-d-2}
%\end{align}
because $\mathcal{E}_{\rm{t}}$ and $\mathcal{E}_{\rm{n}}$ are independent and $\frac{\sigma^2}{||\mathcal{E}_{\rm{n}}||_2^2}$ follows a inverse $\chi^2-$distribution with $D-d$ degrees of freedom and therefore its expectation is $1/(D-d-2)$. Thus, for increasing $D$ while keeping $d$ fixed, Gaussian noise will be increasingly a better approximation for a Gaussian in the normal space. We denote the inflated density with Gaussian noise by $q_{\sigma}(\tilde{x})$ in the following.
%this expected absolute squared error tends to $0$ and 
%More generally, it would be desirable to find constructive and scalable methods for sampling in the normal space. As a first approach, we propose a rejection based method in Section \ref{sec:sampling_in_normal_space}.
%
% by taking the one sample $\varepsilon^*$ out of $k$ which minimizes $\log q_{\sigma}(\tilde{x})$, i.e.
%\begin{equation}\label{eq:}
%\varepsilon^* = \underset{i=1,\dots,k}{\text{argmin}} \log q_{\sigma}(x+\varepsilon_i),
%\end{equation}
%where $\varepsilon_i\sim \mathcal{N}(0,s^2I_D)$ for some suitable $s^2$. Because, s
\ \\ \ \\ 
\textsl{2. }The inflation must not garble the manifold too much. For instance, adding Gaussian noise with magnitude $\sigma\geq r$ to $S^1$ will blur the circle. Since the curvature of the circle is $1/r$, intuitively, we want $\sigma$ to scale with the second derivative of the generating function $f$. Additionally, we do not want to lose the information of $p^*(x)$ by inflating the manifold. If the generating distribution $\pi(z)$ makes a sharp transition at $z_0$, $\pi(z_0-\Delta z_o)\ll \pi(z_0+\Delta z_o)$ for $|\Delta z_o|\ll1$, adding to much noise in $x_0=f(z_0)$ will smooth out that transition. Hence, we want $\sigma$ to inversely scale with $\pi''(z)$. We formalize these intuitions in proposition \ref{prop:nois_choice_JP} and prove it in Appendix \ref{ax:proof_prop}. In accordance with theorem \ref{thm:main}, we say that $p_{\sigma}(\tilde{x})$ approximates well $p^*(x)$ if $\lim_{\sigma\to 0} p_{\sigma}(x) / q_{\rm{n}}(x|x)  = p^*(x)$ for all $x\in \mathcal{X}$ where $q_{\rm{n}}(x|x) $ is the normalization constant of a $(D-d)-$dimensional Gaussian distribution.
\begin{proposition}\label{prop:nois_choice_JP}
	Let $X \in \mathbb{R}^D$ be generated by $U\sim \pi_u(u)$ through an embedding $f:\mathbb{R}^d \to \mathbb{R}^D$, i.e. $f(U)=X$. Let $\pi_u \in C^2(\mathbb{R}^d)$. For $q_{\sigma}(\tilde{x})$ to approximate well $p^*(x)$, in the sense that $\lim_{\sigma \to 0}q_{\sigma}(x)/q_{\rm{n}}(x|x)=p^*(x)$ for $x\in \mathcal{X}$, a necessary condition is that:
	\begin{equation}\label{eq:JP_bound}
	\sigma^2 \ll  \frac{2 \pi_u(u_0)}{||\pi_u''(u_0) \odot (J_f^T(u_0) J_f(u_0))^{-1}||_+}  
%	\left| \frac{\sigma^2}{2 \pi_u(u_0)} ||\pi_u''(u_0) \odot (J_f^T J_f)^{-1}||_+ \right| \ll 1
	\end{equation}
	where $||A||_+=\left| \sum_{i,j=1}^d A_{ij} \right|$  for $A\in \mathbb{R}^{d\times d}$,  and $\odot$ denotes the elementwise product, and $(\pi''(u_0))_{ij}=\frac{\partial^2 \pi_u(u)}{\partial u_i \partial u_j}|_{u=u_0} $ is the Hessian of $\pi_u$ evaluated at $u_0=f^{-1}(x)$.
\end{proposition} 
Intuitively, a second necessary condition is that the noise magnitude should be much smaller than the radius of the curvature of the manifold which directly depends on the second-order derivatives of $f$. This can be illustrated in the following example:
%The proof can be found 
%\footnote{Technically, the circle does not fulfill the conditions of Proposition \ref{prop:nois_choice_JP} since the domain of $f$ is not $\mathbb{R}$.}
\begin{example}\label{exp:JP_bounds_circle}
	For the circle in $\mathbb{R}^2$ generated by $f(u)=(\cos(u),\sin(u))^T$ and a von Mises distribution $\pi_u(u)\propto \exp(\kappa\cos(u))$, we get that $\sigma^2 \ll \min\left(    \left| \frac{2r^2 }{\kappa(\kappa\sin^2(u)-\cos(u))}\right| , r^2 \right)$ where the first condition comes from proposition \ref{prop:nois_choice_JP} and the second one comes from the curvature argument.	
\end{example}
Even though this bound may not be useful as such in practice when $f$ and $\pi_u$ are unknown, it can still be used if $f$ and $\pi_u$ are estimated locally with nearest neighbor statistics. 

From a numerical perspective, inflating a manifold using Gaussian noise circumvents degeneracy problems when training a vanilla NF for low-dimensional manifolds. In particular, the flows Jacobian determinant becomes numerically unstable, see equation (\ref{eq:NF_objective_fct}). This determinant is essentially a volume-changing factor for balls. From a sampling perspective, these volumes can be estimated with the number of samples falling into the ball divided by the total number of points.\footnote{However, if $D$ is large, this is very inefficient due to the curse of dimensionality.} Therefore, we suggest to lower bound $\sigma$ with the average nearest neighbor obtained from the training set to make sure that these volumes are not empty and thus avoid numerical instabilities.
\ \\  
\textsl{3. }Intuitively, if the curvature of the manifold is not too high and if the manifold is not too entangled, $Q-$normal reachability is satisfied for a sufficiently small magnitude of noise. In the manifold learning literature, the entanglement can be measured by the reach number. Informally, the reach number provides a necessary condition on the manifold such that it is learnable through samples, see Chapter 2.3 in \citet{reach_density}. 

{Formally, the reach number is the maximum distance $\tau_{\mathcal{X}}$ such that for all $\tilde{x}$ in a $\tau_{\mathcal{X}}-$neighborhood of $\mathcal{X}$, the projection onto $\mathcal{X}$ is unique. In Appendix \ref{ax:proof_reach} we prove theorem \ref{thm:reach} which states that any closed manifold $\mathcal{X}$ with $\tau_{\mathcal{X}}>0$ is $Q-$normally reachable. }
%, $\tau_{\mathcal{X}}>0$,
\begin{theorem}\label{thm:reach}Let $\mathcal{X}\subset \mathbb{R}^D$ be a closed $d$-dimensional manifold. If $\mathcal{X}$ has a positive reach number $\tau_{\mathcal{X}}$, then $\mathcal{X}$ is $Q-$normally reachable where $Q:=\{q_{\rm{n}}(\cdot|x)\}_{x\in\mathcal{X}}$ is the collection of uniform distributions on a ball with radius $\tau_{\mathcal{X}}$, i.e. $q_{\rm{n}}(\tilde{x}|x)=\mathrm{Uniform}(B(x,\tau_{\mathcal{X}})\cap N_x)$ where $B(x,\tau_{\mathcal{X}})=\{y\in \mathbb{R}^D, \text{s.t. } ||x-y||_2 < \tau_{\mathcal{X}} \}$ denotes the $D-$dimensional ball with radius $\tau_{\mathcal{X}}$ and center $x$.
\end{theorem}
To appreciate theorem \ref{thm:reach}, we refer to the Tubular Neighborhood theorem, which states that every smooth and compact manifold has positive reach (see e.g. \citet{lee_smooth} for a proof).

%% file: JMLR_related_work.tex
\section{Related work}\label{sec:related_work}
Here, we give an overview of methods based on NFs for density estimation on low-dimensional manifolds.
%Recently, a few attempts have been made to overcome the topological constraints of NFs. 
One direction of research concentrates on densities defined on a given manifold, such as spheres, tori or hyperboloids (\Citealp{rezende2015variational},\Citealp{rezende2020normalizing}, \Citealp{mathieu2020riemannian}, \Citealp{lou2020neural}). 
%We follow the notation from \Citealp{brehmer2020flows} and call this approach Flow on Manifold (FOM) in the following. 
Orthogonal to that direction, \Citealp{brehmer2020flows},  \Citealp{beitler2018pie}, \Citealp{kim2020softflow}, \Citealp{cunningham2020normalizing} do not rely on an explicit chart while focusing on improving the generative ability. % of NFs
%The latter works can be further distinguished by their notion of n
%We note that all these approaches assume that the manifold is generated by a single chart.
From the latter works, only \cite{brehmer2020flows} learn, in theory, the density on the manifold $p^*(x)$ exactly. 
%
%\Citealp{bigdeli2020learning} add, as we do, some Gaussian noise to the input data to inflate the manifold. By exploiting some theoretical results of the denoising auto-encoder literature, they estimate the density, up to a constant, while learning to denoise the data. The constant needs to be estimated using Monte-Carlo integration, and thus exactness of the density estimate suffers from the curse of dimensionality whenever $D$ is large.

 \Citealp{cunningham2020normalizing} assume that data live on a noisy, i.e. inflated manifold and propose to learn a stochastic inverse $q(z|\tilde{x})$ of the generator $q(\tilde{x}|z)$. To train the parameters of $q(\tilde{x}|z)$, they rely on variational inference making this approach a special case of a Variational Auto Encoder. Their injective noisy flow improves the sampling quality compared to a baseline NF and, in addition, learns a latent representation. However, by construction, they only learn the inflated distribution $q(\tilde{x})$.

\Citealp{kim2020softflow} follow our methodology closely by inflating the manifold so that a usual NF can be used to learn the inflated density. For each sample $x$, they first draw a value $c$ uniformly on $[0,0.1]$, and then add a sample $\nu$ from $\mathcal{N}(0,c^2I_D)$ to $x$, i.e. $\tilde{x} = x+\nu$. They learn the conditional distribution of the inflated manifold, $q(\tilde{x}|c)$, allowing for sampling on the manifold by setting $c=0$. Their method does not require any knowledge of the manifold (neither the chart, nor the dimensionality), and improve 3D point cloud generation. However, they don't provide a deflation of the inflated distribution, and thus don't learn $p^*(x)$ exactly.

\Citealp{beitler2018pie} propose to use different reference measures for the flow to encode the relevant manifold and irrelevant off-manifold directions. They propose to model the first $d$ latent variables, say $u$, of the flow as standard Gaussian and the remaining $D-d$ variables, say $v$, as a diagonal Gaussian with small variance. The hope is that maximum likelihood training is sufficient to encode the manifold in the first $d$ components, so that a sampling procedure where the remaining $D-d$ components are set to 0, i.e. $v=0$, would produce samples on the manifold. The gist is very similar to our idea expressed in equation (\ref{eq:main_idea}). However, in general, this does not lead to the right density on the manifold, as explained in a footnote on page 4 in \Citealp{brehmer2020flows}, which justifies the name Pseudo-invertible encoder (PIE). Nevertheless, as noted by \Citealp{brehmer2020flows}, it is surprising that "somehow in practice learning dynamics and the inductive bias of the model seem to couple in a way that favor an alignment of the level set $v=0$ with the data manifold. Understanding these dynamics better would be an interesting research goal." Our work gives a theoretical explanation of why the PIE-model favors that alignment: When adding  noise with small magnitude to the dataset (e.g. dequantization for images),  the resulting density can be well approximated by a product of $p^*(x)$ and the noise distribution $q(\tilde{x}|x)$, such that treating the latent variables $u$ and $v$ differently, and thus having a product of two different measures as reference measure, biases the flow to learn this product form. A further interesting future direction would be to make this bias more explicit by constructing a flow for which the Jacobian determinant is in such a product form as well. 

In \Citealp{brehmer2020flows}, the generating chart $f:\mathbb{R}^d\to \mathbb{R}^D$ is learned simultaneously with $p^*(x)$. They first transform $x$ using a usual flow on $\mathbb{R}^D$, and then project to the first $d$ components which is their proposal for $f^{-1}$. They then use another flow to learn the latent density $\pi_u$. To avoid calculating the Gram determinant of $f$, which is computationally expensive especially for $D\gg d$, they propose to train the parameters of $f$ using the mean squared error while updating the parameters of $\pi_u$ using maximum likelihood. They call the former manifold learning phase and the latter density learning phase. Different learning schemes (alternating and sequential) are proposed to ensure that $f$ encodes the manifold and $\pi$ captures the density. For the alternating scheme, they alternate for every epoch between a manifold training phase (updating the parameters of $f$), and the density training phase (updating the parameters for learning $\pi_u$). The experiments conducted by \Citealp{brehmer2020flows} seem to verify that, indeed, $p^*(x)$ is learned exactly. Nevertheless, the ad-hoc training procedures without a unified maximum likelihood objective requires some further experimental verification.\footnote{We further motivate this requirement in Appendix \ref{AX:sphere_related_work}.} 

State-of-the-art methods for image generation based on NF dequantizes the training data as a preprocessing step, see e.g. \Citealp{kingma2018glow}. This dequantization is essentially an inflation of the data-manifold and is typically based on uniform noise. For images, it is generally assumed that $D\gg d$, and thus a dequantization based on Gaussian noise allows us to interpret the dequantization as a thickening of the data-manifold in the normal direction.
%Since their model is the only one capable of learning $p^*(x)$ exactly

%Table \ref{tbl:overview} displays the differences of the models compactly.
%
%
%\begin{center}
%	\begin{table}[H]\label{tbl:2D_von_Mise}
%		\begin{tabular}{l c c c c p{3cm} c|}
%			\hline
%			Model  & Manifold  & Sampling & Latent  & $p^*(x)$& Remark & Ref.  \\ 
%			\hline
%			FOM   & prescribed  &\checkmark & \checkmark & \checkmark & would work for a given atlas  & \Citealp{gemici2016normalizing}, \Citealp{rezende2020normalizing},\Citealp{mathieu2020riemannian} \\
%		
%			$\mathcal{M}$-flow  & learned  &\checkmark & \checkmark & \checkmark & no unified maximum likelihood training, density evaluation may be slow for $D\gg d$, single chart necessary for exactness &  \Citealp{brehmer2020flows} \\
%			
%			
%			
%			INF  & learned &  \checkmark  & \checkmark & \xmark & single chart assumption  & \Citealp{cunningham2020normalizing} \\
%			PIE  & learned & \checkmark   & \checkmark & \xmark & noise needs to be added for manifold data & \Citealp{pie}
%			\\
%			Ours & learned  &\checkmark & \checkmark & \checkmark & $Q$-normal separability condition and $\widetilde{\mathcal{X}}$ diffeomorphic to $\mathbb{R}^D$ assumption for exactness\\
%			\hline
%		\end{tabular}
%		\caption{Comparing estimation methods for densities supported on low-dimensional manifolds, based on NFs.}
%		\label{tbl:overview}
%	\end{table}
%\end{center}

%% file: JMLR_results.tex
\section{Results}\label{results}
We have three goals in this section: first, we numerically confirm the scaling factor in equation (\ref{eq:main_result}) for different manifolds. 
%In fact, we want to show that the scaling factor in equation \ref{eq:main_result} is indeed the right one. 
Second, we verify that Gaussian noise can be used to approximate a Gaussian noise restricted to the normal space. Third, we numerically test the bounds for $\sigma^2$ derived in Section \ref{sec:FG_as_NS}. For training details, we refer to Appendix \ref{ax:de_on_manifolds}. The code for our experiments can be found on \url{https://github.com/chrvt/Inflation-Deflation}.

The \textsl{standard procedure} for our experiments and for evaluating the learned density is the following: % (see Figure \ref{fig:intuition})
\begin{enumerate}[label=\arabic*.]
	\setcounter{enumi}{-1}
	\item \textbf{Data generation: }We sample latent variables $u \sim \pi_u(u)$ for a given $\pi_u(u)$, and generate points $x$ on the manifold using a mapping $f$, i.e. $x=f(u)$.
	\item \textbf{Inflation: }We add noise $\varepsilon$ to $x$, $\tilde{x}=x+\varepsilon$, either in the normal space $N_x$ or in the full ambient space. As an acronym for our inflation-deflation method we use ID. In particular, when the inflation is performed in the normal space, we call the method Normal Inflation Deflation (NID) and when the inflation is isotropic, we call it Isotropic Inflation Deflation (IID).
	\item \textbf{Training: }We learn the inflated distribution, i.e.  $q_{\sigma}(\tilde{x})$ in case of isotropic noise or $q_{\rm{n}}(\tilde{x})$ in case of normal noise, using a Block Neural Autoregressive Flow (BNAF) introduced in \citet{de2019block}.
	\item \textbf{Deflation: }Given an estimator $\hat{q}_{\rm{n}}(\tilde{x})$, we use equation (\ref{eq:main_result}) to calculate $p^*(x)$. For a $d-$dimensional manifold embedded in $\mathbb{R}^D$, the scaling factor when using Gaussian noise is $q_{\rm{n}}(x|x)=(2\pi \sigma^2)^{\frac{d-D}{2}}$.
	\item \label{standard_procedure_KS} \textbf{Quantitative evaluation: }To quantify the quality of the learned density beyond visual similarity, we use the estimate of $p^*(x)$ to approximate $\pi_u(u)$. These densities are related through the Gram determinant of the generating mapping $f$, $\det G_{f}$, see Section \ref{sec:problem} . For that, we calculate the KS statistics between this estimate $\hat{\pi}$ and the ground truth $\pi$. The KS statistic is defined as  %$,$
	\begin{equation}\label{eq:KS} 
	KS = \sup_{u\in \mathcal{U}} | F(u)-G(u) |
	\end{equation}
	where $F$ and $G$ are the cumulative distribution functions associated with the probability densities $\pi_u(u)$ and $\hat{\pi}_u(u)$, respectively. By definition, $KS \in [0,1]$ and $KS=0$ if and only if $\pi_u(u)$ is equal to $\hat{\pi}_u(u)$ for almost every $u\in \mathcal{U}$. Note that, if our estimate does not yield a density on the manifold (i.e. it is not normalized to $1$), the KS statistics still serves as a relative performance measure as the KS value will be lower bounded by a strictly positive number in this case ($1$ minus the corresponding normalization constant).\\
%	the ordering  of the random variables matters. More concretely,
	In 2D,  comparing two random variables based on $\mathbb{P}(X_1\leq x_1,X_2\leq x_2)$ or based on $\mathbb{P}(X_1\leq x_1,X_2\geq x_2)$ (or any of the other two combinations) may lead to different results. Hence, for the KS value in 2D, we need to calculate the KS statistics based on all possible orderings and then take the maximum.\footnote{Note that we are using the KS statistics in a somewhat unusual way. Indeed, the standard KS statistics compares an empirical distribution with an explicit distribution while we compare here the ground truth density $\pi_u$ with the estimated density $\hat{\pi}_u$. The supremum is computed as a maximum over evenly spaced points over $\mathcal{U}$.}
	
	\item \label{standard_procedure_sigma_bounds}  \textbf{$\sigma^2-$bounds: } In proposition \ref{prop:nois_choice_JP}, we derived a necessary condition in form of an upper bound $\sigma_{\text{Prop}}^2$ for $\sigma^2$ such that NID can be well approximated by IID. In addtion, we argued that $\sigma^2$ should not exceed the curvature radius, see example \ref{exp:JP_bounds_circle}. For $d=1$, this curvature radius is straight forward computed using the first and second derivatives of the generating mapping. For $d=2$, we use the Gauss curvature $\sigma_{\text{Gauss}}^2$ instead, see e.g. \citet{do2016differential}. Therefore, as upper bound for $\sigma^2$, we sample $10^4$ points from the target distribution, calculate $\min \{\sigma_{\text{Prop}}^2,\sigma_{\text{Gauss}}^2\}$ for each point, and then take the average. For manifolds with $\sigma_{\text{Gauss}}^2=0$, we only use $\sigma_{\text{Prop}}^2$. As a lower bound $\sigma_{\text{LB}}^2$, we proposed to calculate the $L_2-$distance to the nearest neighbour. Therefore, we sample $10^4$ points from the target distribution, calculate the nearest neighbour for each point, and then take the average
	
	\item \textbf{Benchmarking: }For a known manifold consisting of a single chart $f$, \citet{gemici2016normalizing} used $f$ to encode the manifold into the corresponding latent space $\mathbb{R}^d$, and then learn the latent density using a standard NF. However, manifolds such as spheres or tori, cannot be described using a single chart. In \citet{brehmer2020flows}, such degeneracy problems were avoided numerically by simply moving points that are close to singularities away from them. As a consequence, the density close to these singularities cannot be learned exactly (we illustrate this in the first experiment on $\mathbb{S}^1$). In \citet{brehmer2020flows}, this method is named Flow on manifolds (FOM) and we stick to this notation in the following. In our case, as we are evaluating the qualitive performance using the KS-statistics on the latent densities, we simply train a standard NF directly on the latent space for the remaining experiments (thus avoiding potential degeneracy problems altogether). \footnote{For the case where the latent density is $1$, we use a Gaussian-Kernel density estimator to estimate the latent density. Othewise, we use BNAF.}
\end{enumerate}
\begin{remark} For our qualitative evaluation using the KS-statistics, we rely on being able to relate the density in the data-space $p^*(x)$ with the latent distribution $\pi_u(u)$ via the Gram determinant of the manifold generating mapping $f$. If $f$ consists of singularities, the KS-statistics is still well-defined if these singularities have $0$ measure (as it is the case for e.g. spheres or tori). However, note that our method does not rely on a specific embedding and thus avoids degeneracy problems during training. The IID model does not even need any explicit knowledge of the manifold except its dimensionality for the right scaling factor. We validate the generality of our method by learning a manifold which cannot be described by a single chart covering all the points up to a set of measure $0$. For that, we glue a half sphere with the positive part of a hyperboloid (compactly denoted as $(\mathbb{HS})^2$), see table \ref{tbl:manifolds} .
\end{remark}
\subsection{Proof of concept: $\mathbb{S}^1$}\label{experiment:vonMises}
We start with a circle of radius $3$, a $1-$dimensional manifold embedded in $\mathbb{R}^2$, see table \ref{tbl:circle} . 

\begin{table}[H]
	\centering
	\begin{tabular}{cccc}
		\toprule
		%		\multicolumn{5}{c}{StyleGAN $d=2$ \qquad \qquad \qquad  \qquad \qquad  StyleGAN $d=64$}       \\ 
		%		\cmidrule(r){1-7}
		 feature  & $\mathcal{U}$ & $f(u)$ & $\det G_{f}(x)$    \\
		\midrule
		 closed & $[-\frac{\pi}{2},\frac{\pi}{2}]$ & $
		3 \begin{pmatrix}
		\cos(u) \\
		\sin(u)
		\end{pmatrix}$ & $3$ \\  %& $z_2\in \{0,\pi\} $ \\
		%		$\mathcal{M}-$flow & $4.85 \pm 0.14$ & $309.32 \pm 6$ & $19.95 \pm 0.22$ &  $1019.18 \pm 30.08$ \\
		%		DNF& $\textbf{4.42} \pm 0.2$ &  $\textbf{225.78} \pm 14.4$  & $\mathbf{17.61}\pm 0.11$ & $\mathbf{899.35}\pm 37.19$  \\
		%		InfoMax-VAE& $38.2 \pm 1.19$ &  $12047.92  \pm 105.95$  & $58.18\pm6.67$  & $16397.54 \pm 286.65$  \\
		%		PAE & $22.58 \pm 4.50$ & $7181.21 \pm 27.17$ & $54.51 \pm 7.13$ &  $7169.97 \pm 14.25 $ \\		
		\bottomrule \\
	\end{tabular}
	\caption{Characteristics of $\mathbb{S}^1$.}
	\label{tbl:circle}
\end{table}
We let $\pi_u(u)\propto \exp(8\cos(u))$ be a von Mises distribution. \\ \ \\
%Given $z \sim  \pi_u(u)$, we generate a point in $\mathcal{X}$ according to the mapping $f(u)=3(\cos(u),\sin(u))$. We want to learn the induced density $p^*(x)$. Note that $3p^*(x)=\pi_u(u)$ since $1/3$ is the square root of the Gram determinant of $f^{-1}$. 
%\footnote{The mapping $f$ provides a chart for the entire circle by adjusting the range of $u$ accordingly.}  
%	Therefore, the Riemannian volume form does not depend on $u$ and is given by $dV_G(x)=1/3$.
%To benchmark our performance, we use the idea in \citet{gemici2016normalizing} to first embed the circle into $\mathbb{R}$, using e.g. $f^{-1}$, learn the density there with an NF, and transform this learned density back to $S^1$. 
%In \citet{brehmer2020flows}, this method is named Flow on manifolds (FOM) and we stick to this notation in the following. Note that $f$ is not injective and to illustrate the benefit of our method we choose the singularity point to be $(1,0)^T$. By moving points close to $(1,0)^T$ slightly away from $(1,0)^T$, we numerically ensure that $f$ is an embedding.
\textbf{Inflation and Deflation: }
We inflate $\mathcal{X}$ using 3 types of noise: Gaussian in the normal space (NID), Gaussian in the full ambient space (IID), and $\chi_2$-noise in the normal space as described in example \ref{ex:inflation}(b) with scale parameter $3$. Technically, Gaussian noise violates the $Q-$normal reachability assumption. However, if $\sigma^2$ is small and the scale parameter for the von Mises distribution is large enough, this is practically fulfilled. 
Given an estimator for $q_{\rm{n}}(\tilde{x})$, we use equation (\ref{eq:main_result}) to calculate $p^*(x)$. For the IID and NID methods, we have that $q_{\rm{n}}(x|x)=1 / \sqrt{2\pi \sigma^2}$ and for the normal $\chi^2-$noise is $q_{\rm{n}}(x|x)=\sqrt{3} e^{-3/2}/ (\sqrt{8} \Gamma(\frac{3}{2}))$.
\subsubsection{Full Gaussian vs. Normal space noise}
\begin{figure}[H] \label{fig:normal_1D_vM}
	\centering
	\includegraphics[width=1\linewidth]{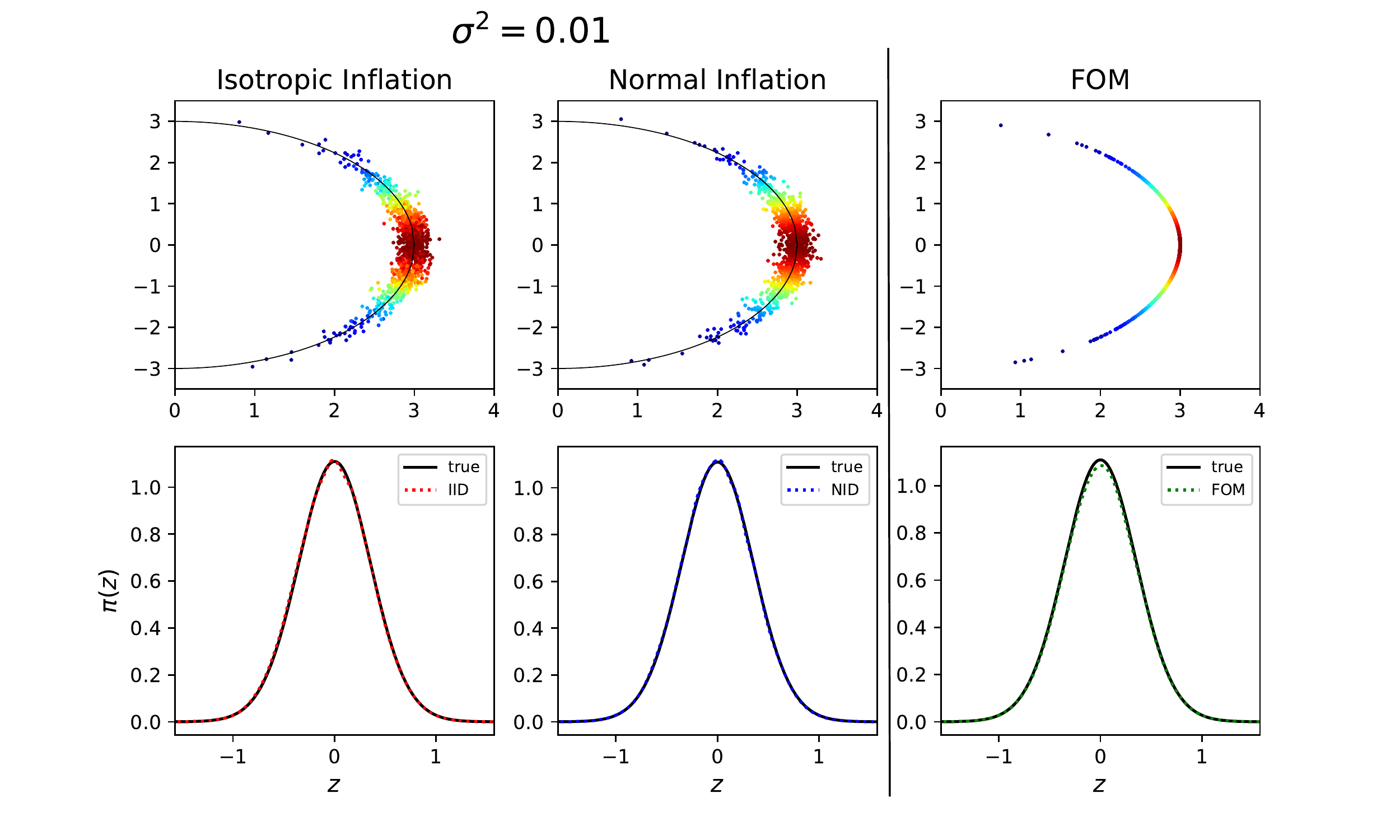}
	\includegraphics[width=1\linewidth]{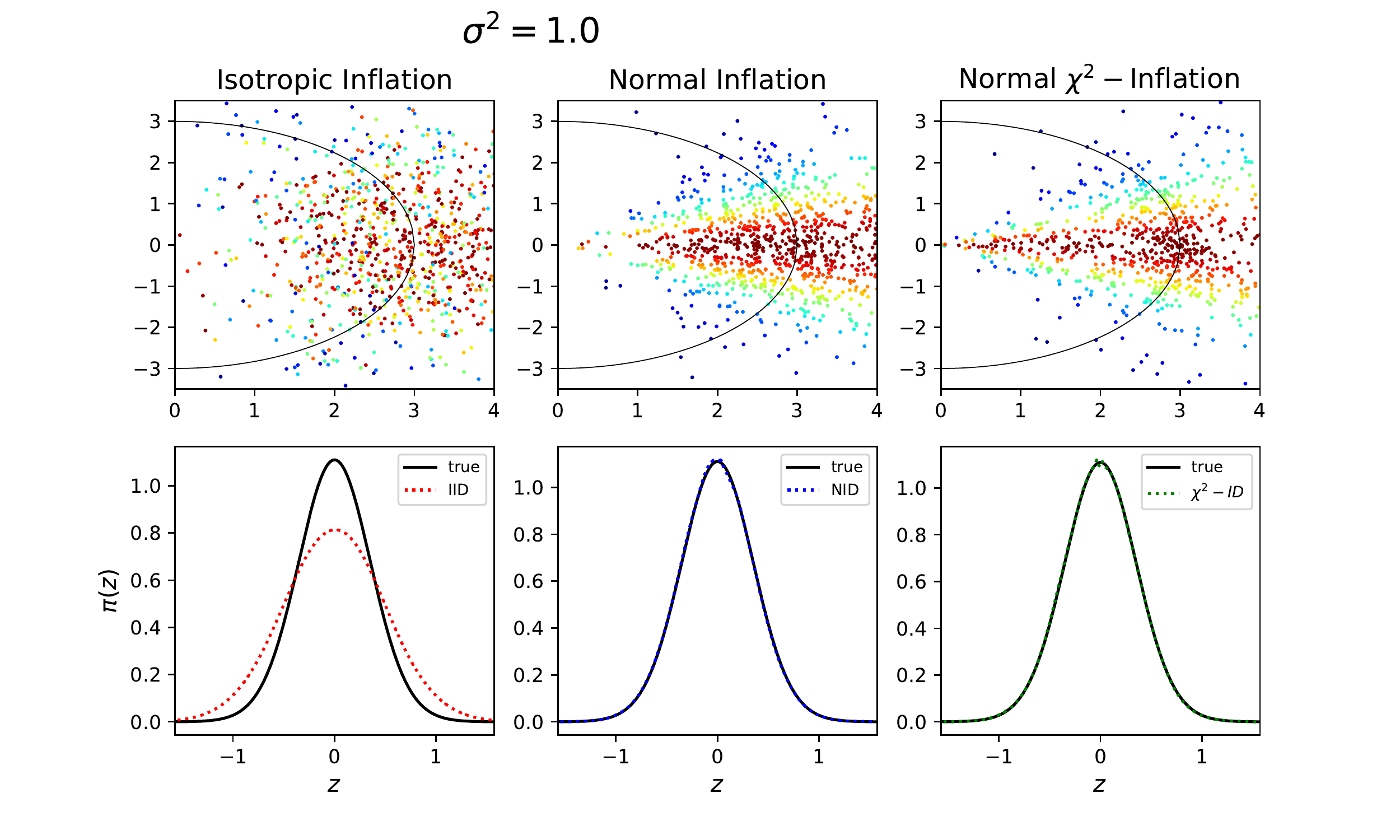}
	\caption{Learned densities for $\sigma^2=0.01$ (above) and $\sigma^2=1$ (below), respectively. \textbf{First row:} Samples used for training the respective model: IID (left), NID (middle), FOM/ $\chi^2$ (right). The black line depicts the manifold $\mathcal{X}$ (a circle with radius 3) and the colors code the value of $p^*(x)$. \textbf{Second row:} Colored line: Learned density $\hat{\pi}_u(u)$  according to equation (\ref{eq:main_result}) multiplied by 3. Blackline: ground truth von Mises distribution. }
	\label{normal_1D_vM}
\end{figure}

In Figure \ref{normal_1D_vM}, we show the results for $\sigma^2 = 0.01$ and $\sigma^2 = 1$. In the respective plot, the first row shows training samples from the inflated distributions $q_{\sigma}(\tilde{x})$ (left), and $q_{\rm{n}}(\tilde{x})$ (middle), respectively. We color code a sample $\tilde{x}=x + \varepsilon$ according to $p^*(x)$ to illustrate the impact of noise on the inflated density. Note that the FOM model (top right) does not need any inflation and therefore is trained on samples from $p^*(x)$ only. In the respective plot, the second row shows the learned density for the different models and compares it to the ground truth von Mises distribution $\pi_u(u)$ depicted in black. As we can see, for $\sigma^2=0.01$ all models perform very well, although the FOM model slightly fails to capture $p(u)$ for $u$ close to $0$ which corresponds to the chosen singularity point (see point 5. in the standard procedure description). For $\sigma^2=1$, we see a significant drop in the performance of the Gaussian model. 
%This is not surprising since for large $\sigma^2$ equation  (\ref{eq:main_idea}) is violated. 
Although the manifold is significantly disturbed, the normal noise model still learns the density almost perfectly \footnote{Note that our method still depends on how well an NF can learn the inflated distribution.}, so does the normal $\chi^2-$noise model, as predicted by theorem \ref{thm:main}. 

\subsubsection{Noise dependence and higher embedding dimensions}
\begin{figure}[h]
	\centering
	\includegraphics[width=1\linewidth]{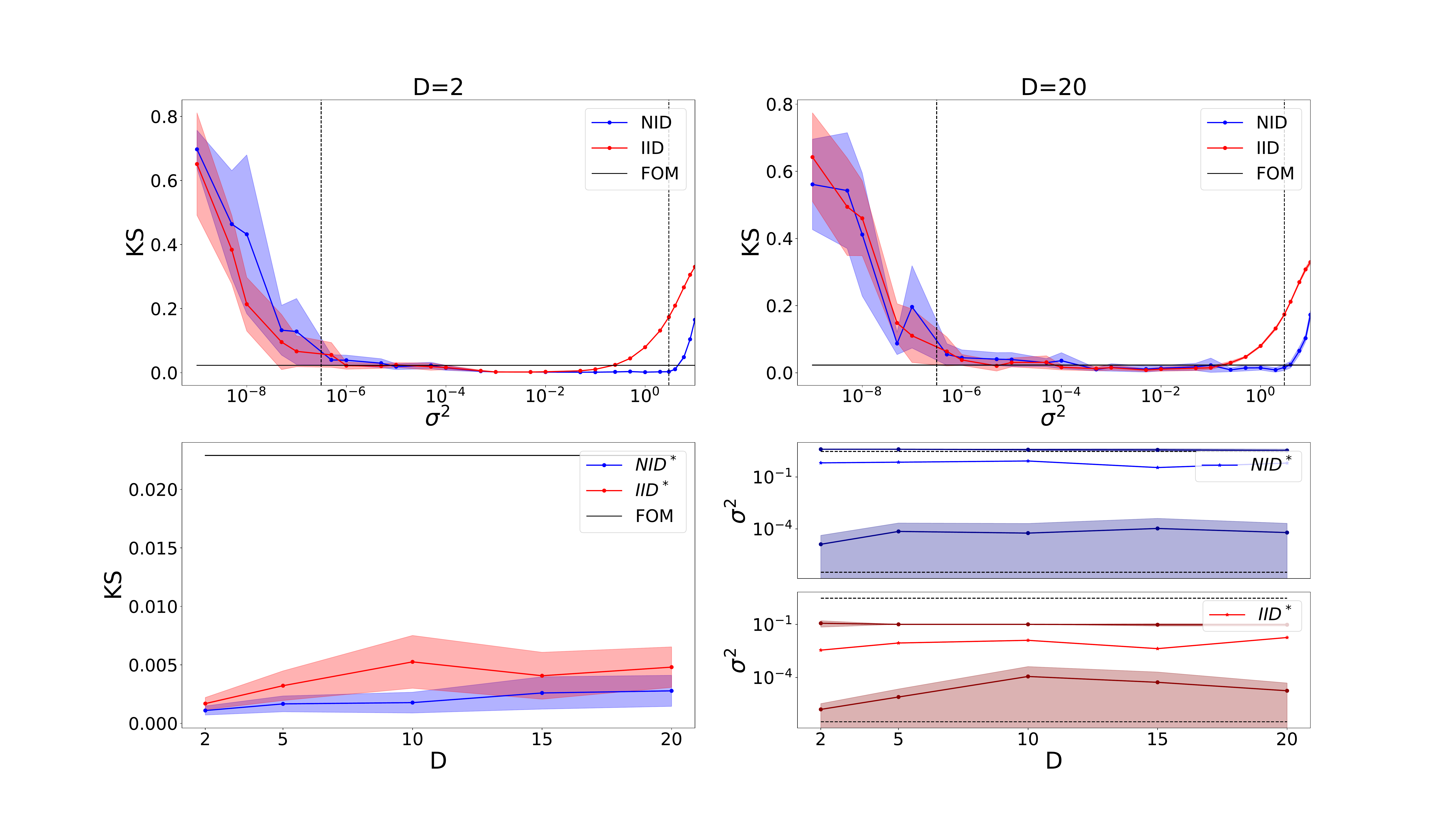} %
	\caption{KS values for the NG- (\textbf{blue}) and IID-noise method (\textbf{orange}) depending on $\sigma^2\in[10^{-9},10]$ and the embedding dimension $D=5,10,15,20$ in log-scale. For $D=2$ (top left) and $D=20$ (top right), the two vertical lines represent the lower and upper bound for $\sigma^2$ estimated according to Chapter \ref{sec:IID_as_NS} with 10K samples. We plot horizontally the KS value obtained from FOM. \textbf{Bottom left:} Optimal KS values depending on $D$. \textbf{Bottom right:} Optimal averaged $\sigma^2$ such that optimal $KS$ is obtained (bright). The maximum and minimum $\sigma^2$ such that the FOM benchmark is outperformed (dark). The dashed horizontal lines are again the theoretical bounds. We used 10 seeds for the error bars and plot in log-scale.}
	\label{vM_KS_statistik}
\end{figure}

To measure the dependence of our method on the magnitude of noise, we iterate this experiment for various values of $\sigma^2$ and estimate the Kolmogorov-Smirnov (KS) statistics. 
%This value allows us to compare two random variables $V$ and $W$ with domain $\mathcal{M}$, or their densities $p$ and $q$, respectively. 
%The KS statistic is defined as $KS = \sup_{x\in \mathcal{X}} | F(x)-G(x) |,$
%%\begin{equation}\label{eq:KS}
%%
%%\end{equation}
%where $F$ and $G$ are the cumulative distribution functions associated with the probability densities $p(x)$ and $q(x)$, respectively. By definition, $KS \in [0,1]$ and $KS=0$ if and only if $p(x)=q(x)$ for almost every $x\in \mathcal{X}$. However, equation (\ref{eq:main_idea}) is only valid if the conditions of theorem \ref{thm:main} are fulfilled. Note that the KS statistics is ill-posed in the case of full Gaussian noise since it doesn't lead to a density on the manifold $\mathcal{X}$.
%%There is no reason why using equation (\ref{eq:main_result}) for the full Gaussian noise would lead to a density on the manifold $\mathcal{X}$. The KS statistics is ill-posed in this case.
% Nevertheless, we are interested in measuring the sensitivity to the noise, and thus consider the KS statistics as a relative performance measure.
In Figure \ref{vM_KS_statistik}, we display the KS values depending on different levels of noise, for the NID (blue) and IID (orange) methods compared with the ground truth von Mises distribution. Also, we embed the circle into higher dimensions $D=5,10,15,20$ and repeat this experiment. The result for $D=2$ and $D=20$ are shown in the first row (left and right).\footnote{Note that the scaling factor depends on $D$, $q_{\rm{n}}(x|x)=1 / (2\pi \sigma^2)^{\frac{D-d}{2}}$.} We add the performance of the FOM model (which is independent of $\sigma^2$) horizontally. Besides, we depict the lower and upper bound for $\sigma^2$ from section \ref{sec:FG_as_NS} with dashed vertical lines.
In the lower-left image, we show the optimal $KS$ values obtained for both models depending on $D$. The lower-right image shows the corresponding $\sigma^2$ for those optimal $KS$. In bright, the optimal average $\sigma^2$ is shown whereas the dark regions are the minimum respectively maximum values for $\sigma^2$ such that we outperformed the FOM benchmark. We note that for both cases, the averaged optimal $\sigma^2$ is within the predicted bounds for $\sigma^2$ (depicted as dashed black horizontal lines). 
%For the NG noise, the upper bound nicely captures the maximal $\sigma^2$ such that we outperform the FOM benchmark.  
%As already seen in Figure \ref{fig:normal_1D_vM}, our method slightly outperforms the FOM model. However, for the Gaussian noise we do not need 
% The upper bound is given by a numerical estimation of equation (\ref{eq:eq:JP_bound}) or example (\ref{exp:JP_bounds_circle}), respectively. Interestingly, it is almost the same as the upper bound stated in Proposition \ref{prop:noise_choice_my}.
%Several aspects are remarkable. 
%The flat course of the KS vs. $\sigma^2$ plot is an  indicator that on $\mathbb{S}^1$ the method is not very sensitive to noise and this does not change with the dimensionality of the embedding space. 
 The optimal KS values do not change much depending on $D$, and the NID and IID models approach each other, as predicted.  

Interestingly, the onset for the increase in the KS value for the NID noise is roughly 3 which is the radius of the circle. For increasing $\sigma^2$, $\tilde{\mathcal{X}}$ resembles more and more a double cone which is not diffeomorphic to $\mathbb{R}^2$ and thus the NF used to train the inflated distribution may not be able to capture the density close to the circle's center correctly. Also, the $Q-$normal reachability is more and more violated with an increasing $\sigma^2$.

\subsection{Densities on surfaces}\label{sec:DE_manifolds}
%$\mathbb{S}^2$}\label{sec:mixture_sphere} % , without using any knowledge about the manifold <---for gaussian noise at least
We show that we can learn different distributions on different manifolds, see table \ref{tbl:manifolds} for an overview of those manifolds and their characteristics.

\begin{table}[H]
	\centering
	\resizebox{\columnwidth}{!}{
	\begin{tabular}{ccccc}
		\toprule
		%		\multicolumn{5}{c}{StyleGAN $d=2$ \qquad \qquad \qquad  \qquad \qquad  StyleGAN $d=64$}       \\ 
		%		\cmidrule(r){1-7}
		\textbf{Manifold} & \textbf{feature}  & $\mathbf{\mathcal{U}}$ & $\mathbf{f(u)}$ & $\mathbf{\det G_{f}(x)}$    \\
		\toprule
	$\mathbb{S}^2$  & closed & $[0,2\pi] \times [0,\pi]$ & $\begin{pmatrix}
		\cos(z_1)\sin(z_2) \\
		\sin(z_1)\sin(z_2) \\
		\cos(z_2)
		\end{pmatrix}$  & $\sin(z_2)$ \\% & $z_2\in \{0,\pi\} $ \\
		\midrule
		$\mathbb{T}^2$ & 	 closed & $[0,2\pi] \times [0,2\pi]$ & $
		\begin{pmatrix}
		(1+0.6\cos(z_2))\cos(z_1) \\
		(1+0.6\cos(z_2))\sin(z_1) \\
		0.6\sin(z_2)
		\end{pmatrix}$  & $0.6(1+0.6\cos(z_2)) $ \\
		\midrule
		$\mathbb{H}^2$  & diffeom. to $\mathbb{R}^2$ & $[0,+\infty) \times [0,2\pi]$ & $
		\begin{pmatrix}
		\sinh(z_1)\cos(z_2) \\
		\sinh(z_1)\sin(z_2) \\
		\cosh(z_1)         
		\end{pmatrix}$  & $(\sinh^2(z_1) + \cosh^2(z_1)) \sinh^2(z_1)$ \\
		\midrule
		thin spiral  & open & $(0,+\infty)$ & $
		3 \pi \sqrt{z}\begin{pmatrix}
		-
		\cos(3 \pi \sqrt{z}) \\
		\sin(3 \pi \sqrt{z})
		\end{pmatrix}$ & $\frac{1+(3 \pi \sqrt{z})^2}{(3 \pi \sqrt{z})^2}$ \\
		\midrule
		swiss roll & open & $(0,1)\times (0,1)$ & $\begin{pmatrix}
		(\alpha+3\pi z_2)\cos(\alpha+3\pi z_2) \\
		21  z_1 \\
		(\alpha+3\pi z_2)\sin(\alpha+3\pi z_2)
		\end{pmatrix}$ & $(63\pi)^2 (1+(0.5+2z_2)^2)$ \\
		\midrule
		$(\mathbb{HS})^2$ & chart 1 & $(-\infty,0] \times [0,2\pi)$ & $
		\begin{pmatrix}
		-\cosh(|z_1|)\cos(z_2) \\
		-\cosh(|z_1|)\sin(z_2) \\
		\sinh(|z_1|)         
		\end{pmatrix}$ & $(\sinh^2(|z_1|) + \cosh^2(|z_1|)) \cosh^2(|z_1|)$ \\
		& chart 2 & $[0,\frac{\pi}{2}] \times [0,2\pi)$ & $\begin{pmatrix}
		\cos(z_2)\cos(z_1+\pi) \\
		\sin(z_2)\cos(z_1+\pi) \\
		\sin(z_1+\pi)
		\end{pmatrix}$ & $\cos(z_1+ \pi)$ \\
%		\bottomrule \\
		%		$\mathcal{M}-$flow & $4.85 \pm 0.14$ & $309.32 \pm 6$ & $19.95 \pm 0.22$ &  $1019.18 \pm 30.08$ \\
		%		DNF& $\textbf{4.42} \pm 0.2$ &  $\textbf{225.78} \pm 14.4$  & $\mathbf{17.61}\pm 0.11$ & $\mathbf{899.35}\pm 37.19$  \\
		%		InfoMax-VAE& $38.2 \pm 1.19$ &  $12047.92  \pm 105.95$  & $58.18\pm6.67$  & $16397.54 \pm 286.65$  \\
		%		PAE & $22.58 \pm 4.50$ & $7181.21 \pm 27.17$ & $54.51 \pm 7.13$ &  $7169.97 \pm 14.25 $ \\		
		\bottomrule 
	\end{tabular}
}
	\caption{Characteristics of various manifolds.}
	\label{tbl:manifolds}
\end{table}

\begin{figure}[H]
	\centering %  width=1\linewidth,height=1\pageheight
	\vspace*{-1cm}
	\includegraphics[width=\textwidth]{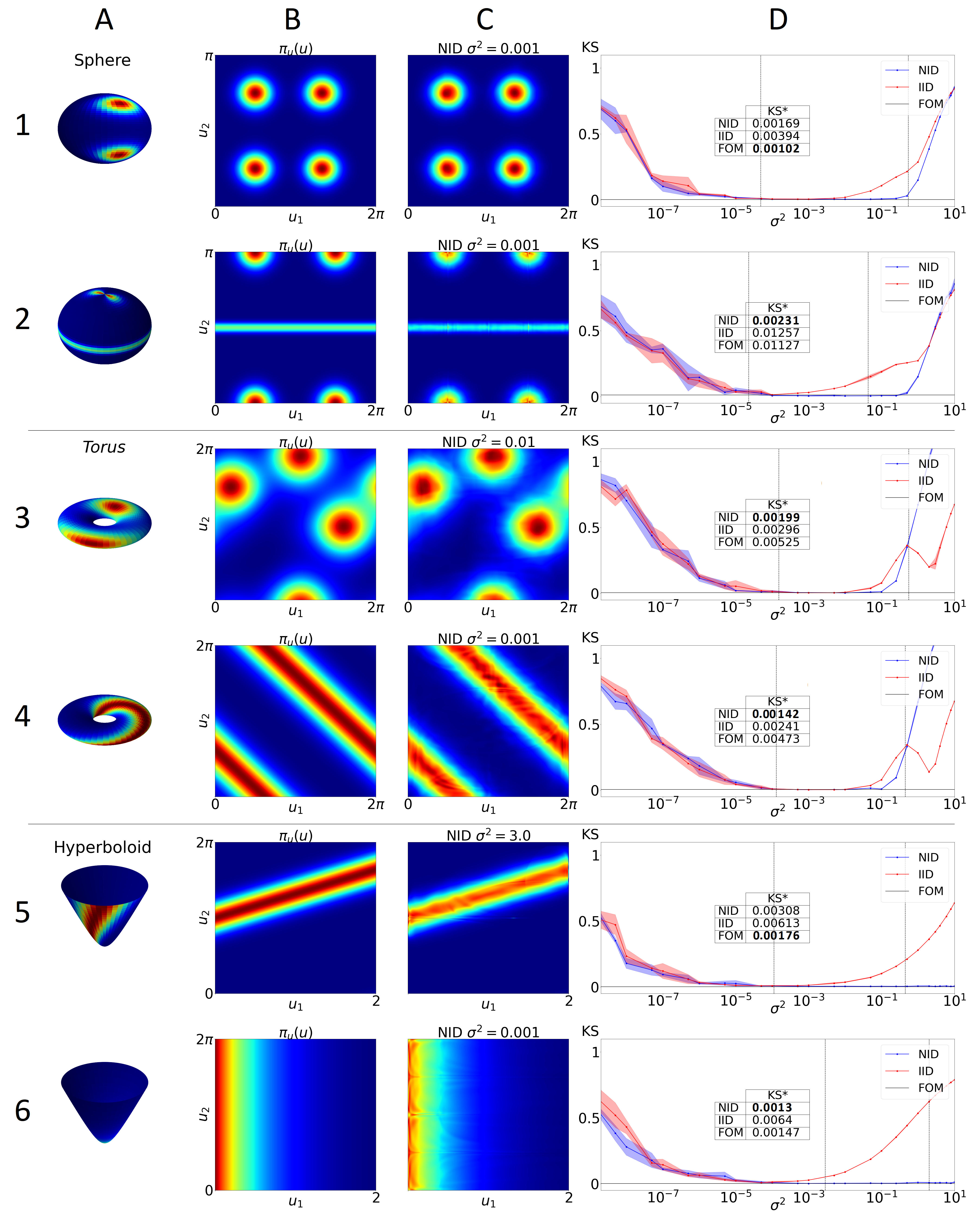} %
%	\vspace*{-2.5cm}
	\caption{\textbf{Columns A and B:} Target density in data (\textsl{A}) and latent space (\textsl{B}) for various manifolds and different latent distributions. \textbf{Column C: }Best learned density using our method with the method NID. \textbf{Column D: }KS vs. $\sigma^2$ plot for the IID and NID methods (we used 3 seeds for the error bars)  with the KS value of FOM as horizontal line. \textbf{Table in D: }Optimal KS values for the different models (best, i.e. lowest, in bold). \textbf{Vertical Lines in C }Lower and upper bound (see point \ref{standard_procedure_sigma_bounds} of the standard procedure). }
	\label{fig:DEmanifolds0}
\end{figure}

In Figure \ref{fig:DEmanifolds0}, we show different target densities in data and latent space (columns A and B), along with the learned latent distributions using our method (as described in point \ref{standard_procedure_KS} of the \textsl{standard procedure}) with the normal inflation-deflation method NID (column C). We take the model with $\sigma^2$ corresponding to the best KS value. In the last column D, we show how the KS-statistics depends on $\sigma^2$ using IID, NID, and the FOM baseline. We refer to Appendix \ref{ax:de_on_manifolds}, \ref{ax:latent_densities} and \ref{ax:additional_figure} for the training details, exact latent densities and additional figures showing the learned latent densities using IID and FOM.

Remarkable, our method performs well on a wide range of manifolds and different target distributions. Whether the manifold is closed (A 1-2, 3-4), open (A 5-6, 7-9), or consists of multiple charts (A 10-11), whether the latent variables are idependent (B 1,3,6,8,10) or dependent (B 2,4,5,9,11), whether the distribution is supported on points for which the Gram determinant is $0$ (A 1-2) or on points for which the Gram determinant is arbitrarily large (A 7), the induced latent density (and therefore the data-density $p^*(x)$) is approximated well. This is not only reflected in the visual similarity to the target distribution (columns B vs. C), but also in the KS statistics (column D). Surprisingly, the best KS values for the IID and NID methods are of the same order as the FOM baseline (tables in D). This is striking as the IID and NID methods are trained in data space, in contrast to the FOM which is trained in latent space directly (see point \ref{standard_procedure_KS} in the \textsl{standard procedure}). In some cases, the NID even outperforms the FOM significantly  (see tables in D 2-3). The optimal KS value for IID is only slightly worse than the one for NID showing that indeed our method can even be used without any explicit knowledge of the manifold (except its dimensionality for the right scaling factor). \\
Note that the NID method always allows for a greater range of $\sigma^2$ compared to IID, except for the thin spiral for which both curves have almost the same course (D 7). As an extreme case, the geometry of the hyperboloid $\mathbb{H}^2$ even  allows for very large values of $\sigma^2$ when using NID (D 5-6). Notably, almost all the KS curves are U-shaped. However, for the torus (D 3-4) and swiss roll (D 8-9) the KS value for IID decreases approaching $\sigma^2=10^1$ before increasing again. For increasing $\sigma^2$, the induced latent distribution $\hat{\pi}_u$ is increasingly flat. Then, certain values of $\sigma^2$ lead to the right scaling such that $\int_{\mathcal{U}} \hat{\pi}_u(u) du \approx 1$ which decreases the KS value. 

Except for the Hyperboloid and thin spiral, D 6 and D 7, respectively, does the lower bound based on nearest neighbor statistics nicely predicts the magnitude of noise required to approximate $p^*(x)$ well using IID. Also the upper bound (which is $>10$ for the Swiss Roll in both cases, see D 8-9, and thus not seen in those figures) behaves as predicted. It is necessary (though not sufficient, see D 5-6) for $\sigma^2$ to be lower than this upper bound such that IID approximates NID well and thus can be used to approximate $p^*(x)$.

%Finally, note that the range for $\sigma^2$ when using IID is significantly smaller for densities concentrated on points $x$ with $\det G_f(x)=0$. For instance, the second  density on the sphere or both densities on the Hyperboloid are all concentrated on such points (see $\det G_f$ in Table \ref{tbl:manifolds}). Arguably, adding too much of full Gaussian noise blurs the density at those points leading to a greater $\sigma^2-$sensitivity.
%
% Remarkable, although we train in data space, the KS value is of the same order as the FOM baseline which is trained in latent space directly (see point \ref{standard_procedure_KS} in the \textsl{standard procedure}). \\ 
%% For some combinations of manifolds ($\mathbb{S}^2$,$\mathbb{T}^2$,$\mathbb{H}^2$, Swiss Roll) and latent distributions, we even outperform the FOM baseline.
%. One possible explanation is that the NF used to learn the inflated distribution is not expressive enough to fully capture such non-linearities. Also different to the other manifolds, the inflation requires more noise ($>10^{-3}$) to  approach the FOM baseline. \\
%The optimal KS value for IID is only slighly worse than the one for NID showing that indeed our method can even be used without any explicit knowledge of the manifold (excepts its dimensionality for the right scaling factor).\\
%Notably, almost all the KS curves are U-shaped. \\

\begin{figure}[H] %\ContinuedFloat
%	\1renewcommand\figurename{\textbf{Figure 5 continuation.}}
	\centering %  width=1\linewidth,height=1\pageheight
	\vspace*{-1cm}
	\includegraphics[width=\textwidth]{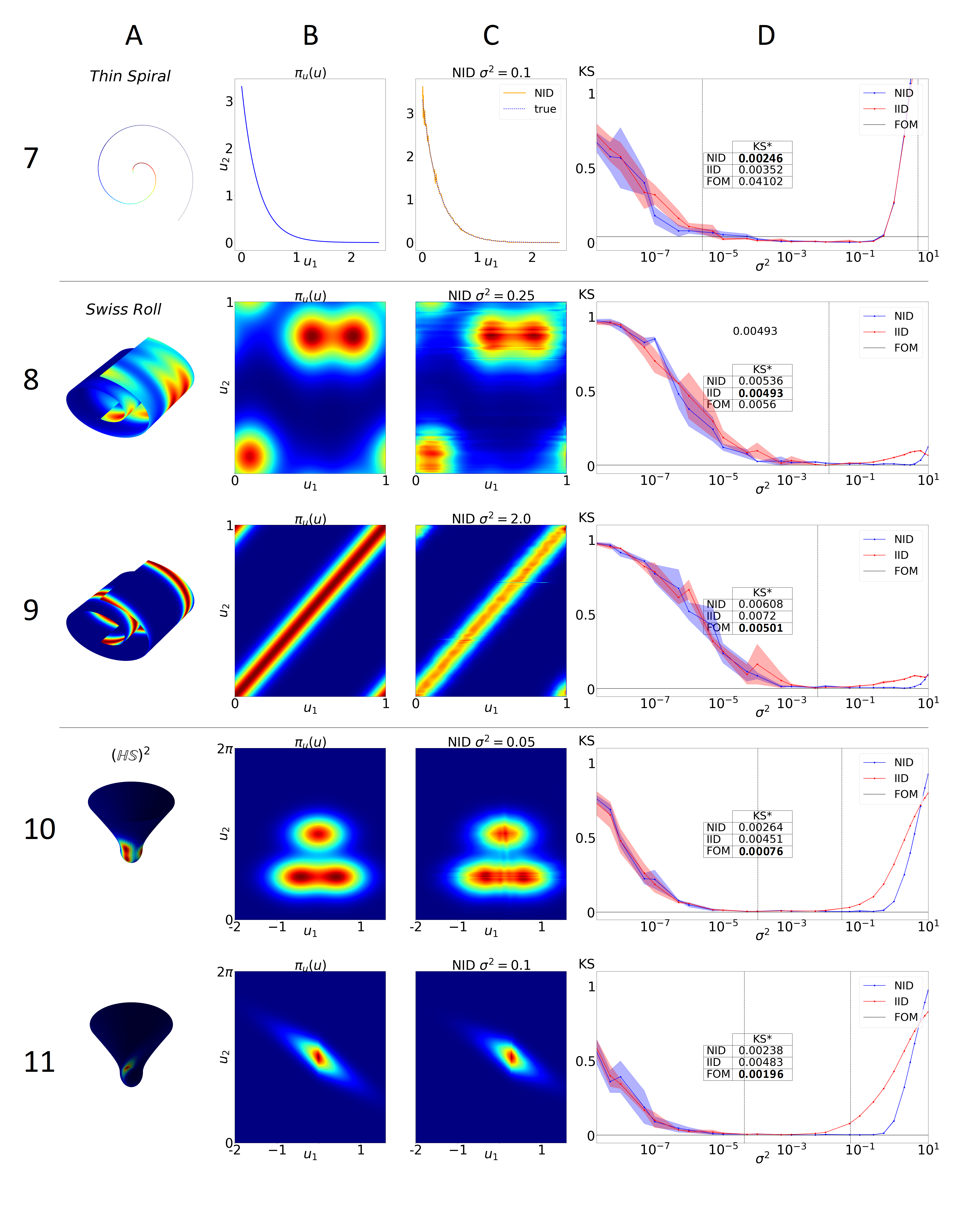}
%	\vspace*{-2cm}
	\captionsetup{labelformat=empty}
	\caption{{Figure 5 (continuation):} \textbf{Columns A and B:} Target density in data (\textsl{A}) and latent space (\textsl{B}) for various manifolds and different latent distributions. \textbf{Column C: }Best learned density using our method with the method NID. \textbf{Column D: }KS vs. $\sigma^2$ plot for the IID and NID methods (we used 3 seeds for the error bars)  with the KS value of FOM as horizontal line. \textbf{Table in D: }Optimal KS values for the different models (best, i.e. lowest, in bold). \textbf{Vertical Lines in C }Lower and upper bound (see point \ref{standard_procedure_sigma_bounds} of the standard procedure). }
	\label{fig:DEmanifolds1}
\end{figure}

\subsection{Density estimation on $SO(2)$}\label{sec:DE_manifolds}
Finally, we consider the manifold consisting of orthonormal $k-$frames in $\mathbb{R}^n$, 
\begin{equation}\label{eq:}
\mathbb{S}^{n,k}:= \left\lbrace V:=[v_1,\dots,v_k] \in \mathbb{R}^{n\times k} \ s.t. \ V^T V = 1\right\rbrace, %_{k\times k} 
\end{equation}
the Stiefel manifold. The dimension of this manifold is $nk-(k+1)k/2$. We consider the special case where $n=k=2$ in the following. Then, $\mathbb{S}^{2,2}$ is the orthogonal group $SO(2)$ with $\dim (\mathbb{S}^{2,2}) = 1$. We can represent an element in $SO(2)$ as $2$ orthogonal vectors on the unit circle embedded in $\mathbb{R}^2$. Therefore, as latent distribution $\pi_u$, we consider a mixture of 1-dimensional von Mises distributions $\pi_u$, and sample an element in $[x,x_{\bot}]\in SO(2)$ as follows:
\begin{enumerate}[label=\arabic*.]
	\item we sample $u\sim \pi_u(u)$,
	\item set $x=(\cos(u),\sin(u))^T$,
	\item rotate $x$ and set $x_{\bot}=(-\sin(u),\cos(u))^T$.
%	\item  following point \ref{standard_procedure_KS} of the \textsl{standard procedure}
\end{enumerate}
Finally, we concatenate $x$ and $x_{\bot}$ to a single vector in $\mathbb{R}^4$ and learn the corresponding density using an NF, see Appendix \ref{ax:de_on_manifolds} and \ref{ax:latent_densities} for the training details and exact latent densities, respectively.

\begin{figure}[h]
	\centering
	\includegraphics[width=\textwidth]{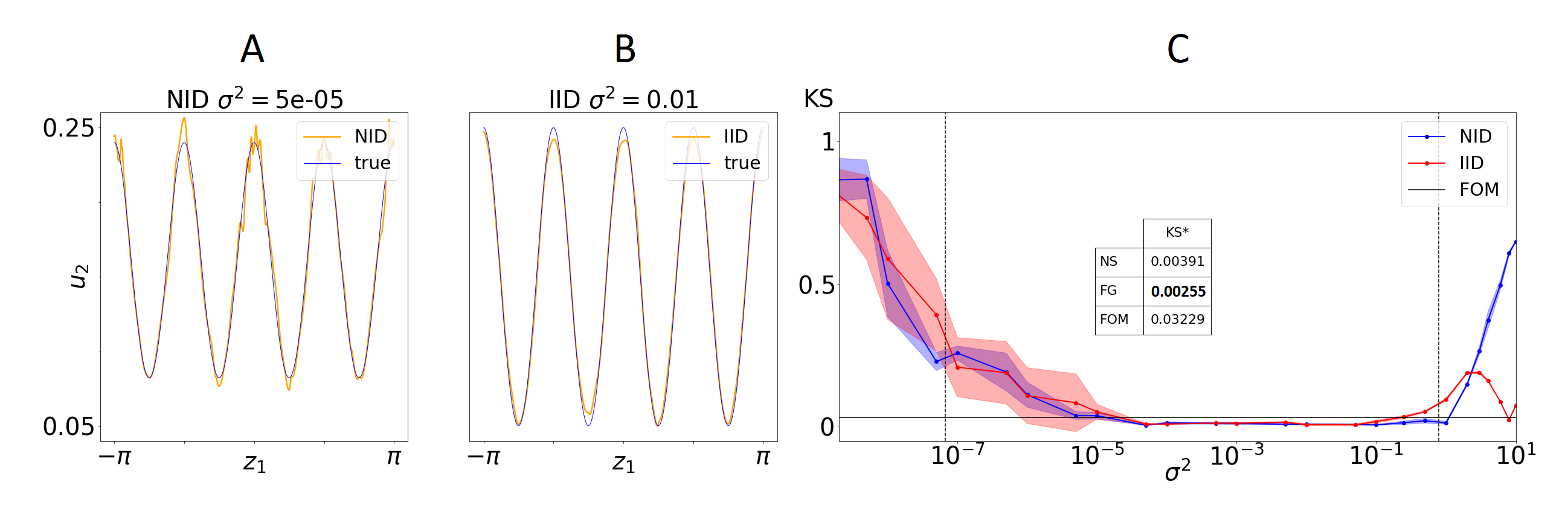} %pdf
	\caption{\textbf{A: } Learned latent density using NID (\textsl{orange}) on top of target latent density (\textsl{blue}). \textbf{B: }Learned latent density using IID (\textsl{orange}) on top of target latent density (\textsl{blue}).   \textbf{C: } KS vs. $\sigma^2$ plot for IID and NID (we used 3 seeds for the error bars)  with the KS value of FOM as horizontal line. \textbf{Table in C: }Optimal KS values for the different models (best, i.e. lowest, in bold). \textbf{Vertical Lines in C: }Lower and upper bound (see point \ref{standard_procedure_sigma_bounds} of the standard procedure).}
	\label{fig:DEstiefel}
\end{figure}

In Figure \ref{fig:DEstiefel}, we plot the induced latent distributions (see point \ref{standard_procedure_KS} of the \textsl{standard procedure}) using NID and IID (with the $\sigma^2$ corresponding to the best KS value) on top of the true latent distribution (Figures A and B). In Figure C, we show how the KS-statistics depends on $\sigma^2$ using IID, NID, and the FOM baseline.

Both, the NID and IID match the target distribution except at the peaks and some valleys of $\pi_u$. The estimate using IID is smoother, however, note that the KS-statistics does not account for the smoothness of the estimate. Indeed, an estimate using NID with e.g. $\sigma^2=0.01$ yields a similar smooth curve as the IID (see Figure \ref{fig:DEstiefel001} in Appendix \ref{ax:additional_figure}).\\
Surprisingly, the optimal KS value for IID is slightly better than the one for NID, both outperforming FOM (using a simple Gaussian-Kernel Density Estimate). However, NID always allows for a greater range of $\sigma^2$ compared to IID, as expected. 
% Hence, in some cases, the KS-statistics can be misleading.
%Suprisingly, the KS value for IID decreases again approaching $\sigma^2=10$.

\subsection{Density estimation on MNIST}\label{sec:DE_MNIST}
Finally, we end this section with an application on the handwritten digit dataset MNIST, \citet{726791}. The manifold hypothesis states that real-world data, such as images, can be described by a few key features only, thus populating a low-dimensional manifold in the high-dimensional embedding space. 

To estimate the density of e.g. digit 1 images, both the inflation-deflation method and the $\mathcal{M}-$flow need to know the manifold dimensionality $d$.\footnote{Note that the inflation-deflation method does only need to know the dimensionality $d$ for the right scaling factor during testing. The $\mathcal{M}-$flow, however, needs to know $d$ for training. In exchange, the $\mathcal{M}-$flow learns a low-dimensional representation which the inflation-deflation method does not.} Estimating this intrinsic dimensionality $d$ is an active research area, see e.g. \citet{10.1145/1102351.1102388} and \citet{facco2017estimating}. For instance, \citet{10.1145/1102351.1102388} estimate the intrinsic dimensionality of MNIST digit 1 to be roughly $8$.
%To estimate $p^*(x)$ both our method and the $\mathcal{M}-$flow need to estimate 

\begin{figure}[h]
 	\centering					
 	\includegraphics[scale=0.25]{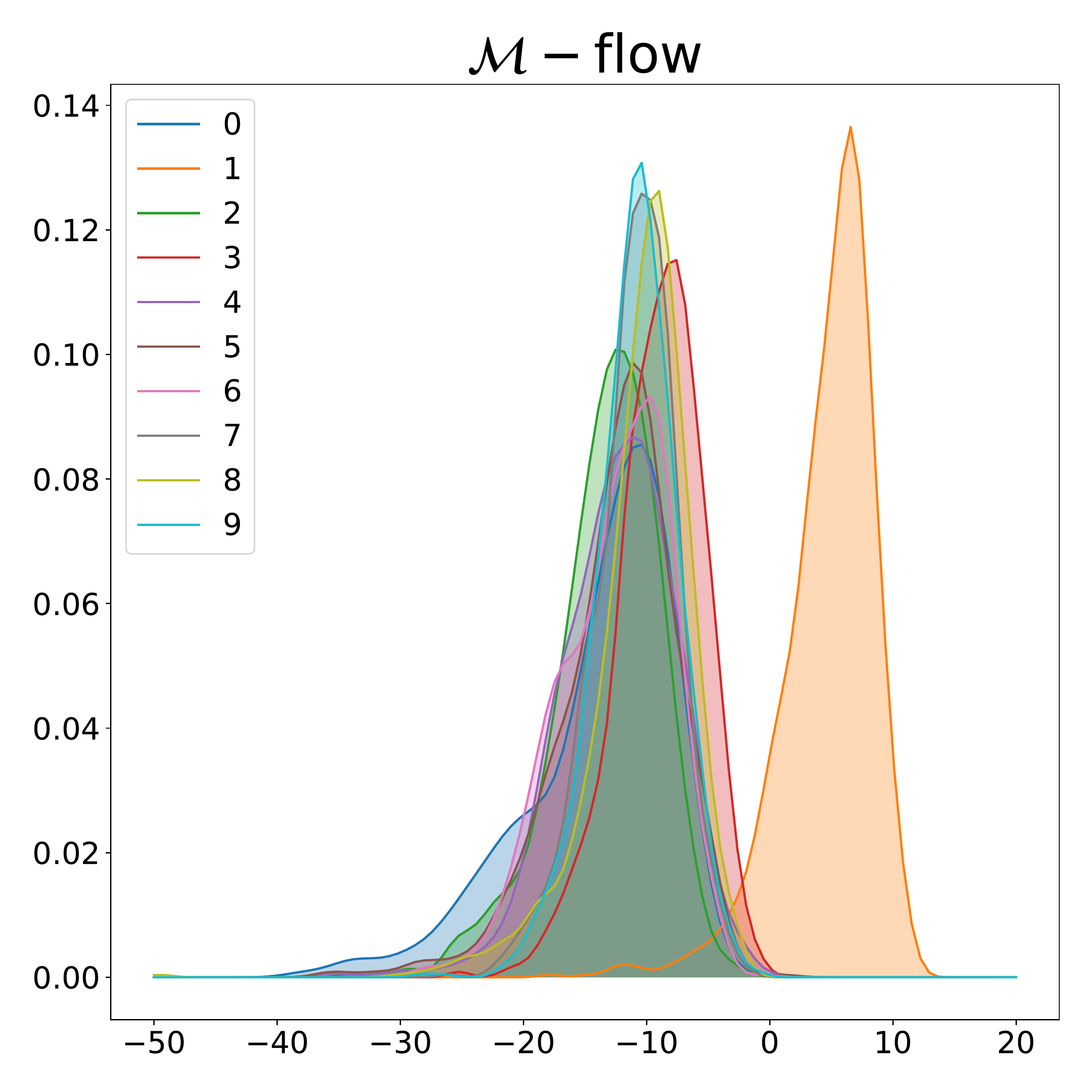}
 	\includegraphics[scale=0.25]{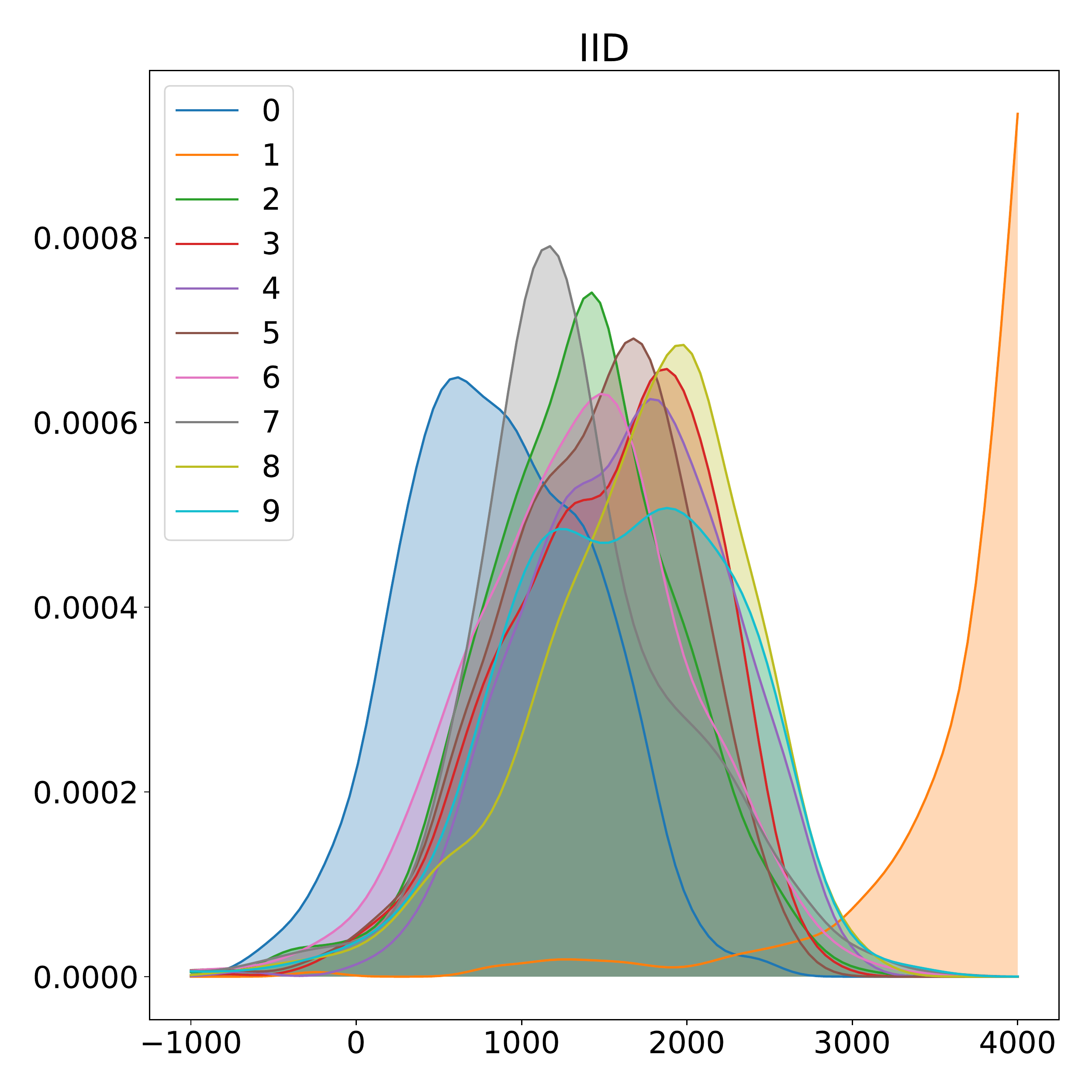}
 	\caption{Log likelihoods on various MNIST test digits using $\mathcal{M}-$flow (\textbf{left}) and IID (\textbf{right}) trained on digit 1 only. }
 	\label{fig:MNIST}
 \end{figure}

We test the utility of learned digit 1 likelihoods for out of distribution detection (OOD) using IID (isotropic inflation-deflation) and the $\mathcal{M}-$flow. In Figure \ref{fig:MNIST}, we show the log-likelihood densities (estimated using kernel density estimation) on the MNIST test set after training on digit 1 images from the training set only. For the IID, we preprocess the training set by adding Gaussian noise with $\sigma^2=0.1$ to the 8-bit images.\footnote{Note that this is different to the usual uniform dequantization performed on images.} For the $\mathcal{M}-$flow, we leave the training set unaltered. Though, we did not find this preprocessing (or the absence of it) to have a significant impact on the log-likelihoods for both methods. We refer to Appendix \ref{ax:de_on_MNIST} for more training details and additional plots for different preprocessing protocols.

In Figure \ref{fig:MNIST}, we want to highlight two interesting observations. First, the log-likelihoods of digits other than $1$ are not significantly different using the IID or $\mathcal{M}-$flow method for OOD. One can see this by comparing the area of intersection of the digit $1$ density (orange) with the other digits (other colors). The greater this area, the more out of distribution examples (in this case MNIST digits other than 1) would be classified as digit $1$ when using a naive classifier based on an adhoc log-likelihood threshold. This area is $\approx 0.07$ for both methods. Our second observation is that the absolute log-likehood values differ substantially. As both method try to esitmate the density $p^*(x)$ supported on a low-dimensional manifold, we would have expected similar log-likelihood values. The fact that these values are several magnitudes apart, together with the observation that an inflation is not strictly necessary using an NF to learn the data-density (see Figure \ref{fig:MNIST_dequatization_flip} in Section \ref{ax:additional_figure}), indicate that the MNIST digit 1 images do not strictly lie on a low-dimensional manifold embedded in $\mathbb{R}^{D}, D=784$. In such a case, the $\mathcal{M}-$flow would still try to fit the training set onto a manifold which may lead to overfitting and unforeseeable log-likelihood values when evaluating on a test set. In contrast, the IID method is by construction robust to overfitting as the addition of isotropic noise leads to similar log-likelihood values in the vicinity of the data-manifold. 
%
%
% The IID is by construction robust to overfitting as 
 
Finally, we want to revisit our remark on the computational complexity of the $\mathcal{M}-$flow, see Section \ref{sec:problem} . To evaluate the density using the $\mathcal{M}-$flow, one needs to calculate the Gram determinant which has a computational complexity of $\mathcal{O}(d^2D)+\mathcal{O}(d^3)$.\footnote{The necessary Jacobian is computed using automatic differentiation.} Indeed, to evaluate 1000 digits using a batch size of 1, the $\mathcal{M}-$flow needs about $17.5$ hours. For the same amount, the inflation-deflation methods needs less than 10 seconds. 
 
%  For that, we train an NF on MNIST digit 1 using different dequantization steps (Uniform, Gaussian with $\sigma^2=1$, Gaussian with $\sigma^2=0.1$) and plot the corresponding log-likehood histograms on the test set for each digit. In Figure 	\ref{fig:MNIST}, 

%Usually, a dequantization step on the training set is performed before training an NF. A dequantization step is essentially an inflation of the data manifold using, typically, uniform noise $\text{Uniform}(0,1)$ on an $8-$bit image $x$.
%Contrary, our theory suggests adding Gaussian noise to ensure that the $Q-$normal reachability condition is fullfilled and the on-manifold density can be learned exactly.

% We test this prediction by comparing the utility of digit 1 likelihoods for out of distribution detection. For that, we train an NF on MNIST digit 1 using different dequantization steps (Uniform, Gaussian with $\sigma^2=1$, Gaussian with $\sigma^2=0.1$) and plot the corresponding log-likehood histograms on the test set for each digit. In Figure 	\ref{fig:MNIST}, 

%\begin{figure}[h]
%	\centering
%	\includegraphics[scale=0.2]{gaussian_histograms.pdf}
%	\includegraphics[scale=0.2]{gaussian_small_histograms.pdf}
%	\includegraphics[scale=0.2]{uniform_histograms.pdf}
%	\includegraphics[scale=0.2]{no_deq_histograms.pdf}
%	\includegraphics[scale=0.2]{gaussian_001_histograms.pdf}
%	\caption{Cool!}
%	\label{fig:MNIST}
%\end{figure}

%% file: JMLR_discussion.tex
\section{Discussion}\label{summary}
To overcome the limitations of NFs to learn a density $p^*(x)$ defined on a low-dimensional manifold, we proposed to embed the manifold into the ambient space such that it becomes diffeomorphic to $\mathbb{R}^D$, learn this inflated density using an NF, and, finally, deflate the inflated density according to theorem \ref{thm:main}. There, we provided sufficient conditions on the choice of inflation such that we can compute $p^*(x)$ exactly.
% Notably, we don't need to explicetly assume that $p^*(x)$ is supported on a flat manifold, i.e. describable by a single chart, such that . 
Our method depends on some critical points which we addressed in Section \ref{sec:FG_as_NS}. So far, the magnitude of noise $\sigma^2$ when using NFs on real-world data is somewhat chosen arbitrarily. As a step to overcome this arbitrariness, we derived an upper bound for $\sigma^2$ in proposition \ref{prop:nois_choice_JP} and established an interesting connection to the manifold learning  literature in theorem \ref{thm:reach} . However, proposition \ref{prop:nois_choice_JP} may not be very useful as such in real-world application and numerical methods need to be considered which potentially suffer from the curse of dimensionality. On a more positive note, our various experiments on different manifolds suggest that a great range for $\sigma^2$ lead to good results, even when using full Gaussian noise. Thus, including $\sigma^2$ into the standard hyperparameter search will likely suffice.\\
Our theoretical results open new research avenues. Using full Gaussian noise to learn the inflated distribution smears information on $p^*(x)$, in particular, if $p^*(x)$ has many local extrema. This loss of information may be especially impactful in out of distribution (OOD) detection or when it comes to adversarial robustness. Therefore, developing methods that allow generating noise in the manifold's normal space could improve the performance of NFs on such tasks. \\
Another interesting direction is to exploit the product form of equation (\ref{eq:main_idea}) and learn low-dimensional representations by forcing the NF to be noise insensitive in the first $d$-components and noise sensitive in the remaining ones. Inverting the corresponding flow allows sampling directly on the manifold.

%% file: JMLR_appendix.tex
\section{Appendix}

\subsection{Proof of theorem \ref{thm:main}} \label{ax:proof_thm}
Let $x\in \mathcal{X}$. Since $\mathcal{X}$ is a $d-$dimensional $C^2$ manifold, there exists an open neighborhood $\mathcal{B}_x$ of $x$ in $\mathcal{X}$, an open set $\mathcal{U}_x$ in $\mathbb{R}^d$, and an invertible map $f:\mathcal{U}_x \mapsto \mathcal{B}_x$, $\mathcal{U}_x\subset \mathbb{R}^d$,  such that $f$ and $f^{-1}$ are twice continuously differentiable. It follows that the Gram determinant of $f$ is non-zero for all $x\in \mathcal{B}_x$, i.e. $\det G_f(x) \neq 0 \ \forall x\in \mathcal{B}_x$.  We exploit this by constructing a local diffeomorphism $\tilde{f}$ on the inflated space $\widetilde{\mathcal{X}} = \bigcup_{x\in \mathcal{X}}N_{q_{\rm{n}}(\cdot|x)}$ in the following. \\
For that, we denote by $A_u$ the matrix with columns consisting of normal vectors spanning the normal space in $x=f(u), u\in \mathcal{U}_x$. Without loss of generality we can set $\det A_u^T A_u=1$. With $\mathcal{V}_x\subset \mathbb{R}^{D-d}$, we define $\tilde{f}:\mathcal{U}_x\times \mathcal{V}_x \subset \mathbb{R}^d \times \mathbb{R}^{D-d} \to \widetilde{\mathcal{B}}_x$ for some $\widetilde{\mathcal{B}}_x \subset \widetilde{\mathcal{X}}$ as follows:
\begin{equation}\label{eq:def_phi}
\tilde{f}(u,v) = f(u) + A_u v.
\end{equation}
Note that, by assumption, $0\in \mathcal{V}_x$. Thus, for $v$ sufficiently small, i.e. $||v||<\varepsilon$ for some $\varepsilon>0$, $\tilde{f}$ is indeed a diffeomorphism which follows from the inverse function theorem.\footnote{In fact, we need to show that $\det J_{\tilde{f}}(u,0)\neq 0$ for all $(u,0)$. Because this implies the existence of a local neighborhood such that $\tilde{f}$ is diffeomorphic to the image of this local neighborhood. That $\det J_{\tilde{f}}(u,0)\neq 0$ follows immediately from Lemma \ref{lemma:det_f} .} Our key observation is that
\begin{equation}\label{eq:gram_relation}
\det G_{\tilde{f}}(x) = \det G_f(x)
\end{equation} 
which allows us to relate the density on $\widetilde{\mathcal{X}}$ to the density on $\mathcal{X}$. For the sake of clarity, we prove equation (\ref{eq:gram_relation}) in Lemma \ref{lemma:det_f} below.

Now let $\tilde{x}=x+\varepsilon_{\rm{n}} \in \widetilde{\mathcal{X}}$ such that $\tilde{x}\in \widetilde{\mathcal{B}}_x$.
Since $\mathcal{X}$ is $Q-$normally reachable, $\mathcal{P}_{\tilde{X}}-$almost all $\tilde{x}$ are uniquely determined by some $(u,v)^T=\tilde{f}^{-1}(\tilde{x})\in \mathcal{U}_x\times \mathcal{V}_x$, and since $u$ and $v$ are sampled independently by assumption, it must hold that
\begin{equation}\label{eq:}
q_{\rm{n}}(\tilde{x})  = (\det G_{\tilde{f}}(\tilde{x}))^{-\frac12} \pi_u(u) \pi_v(v)
\end{equation}
where $\pi_v(v)$ is the noise generating latent distribution. Note that $q_{\rm{n}}(\tilde{x})$ is the density of $d\mathbb{P}_{\tilde{X}}$ with respect to the volume form $dV_{\tilde{f}}$. For $\tilde{x}=x$, we have that $v=0$ and thus 
\begin{equation}\label{eq:}
q_{\rm{n}}(x) = (\det G_{\tilde{f}}(x))^{-\frac12}\pi_u(u) \pi_v(0).
\end{equation}
Now since $\det G_{\tilde{f}}(x) = \det G_{{f}}(x)$, we have that 
%\begin{equation}\label{eq:gram_det_to_show}
%\det G_{\tilde{f}}(x) = \det G_{{f}}(x),
%\end{equation}
%we have that 
\begin{align}\label{eq:}
q_{\rm{n}}(x)  =& (\det G_{\tilde{f}}(x))^{-\frac12} \pi_u(u) \pi_v(0) \notag \\
=& (\det G_{f}(x))^{-\frac12} \pi_u(u) \pi_v(0) \notag \\
=& p^*(x) \pi_v(0) \notag \\
=& p^*(x) q_{\rm{n}}(x|x)
\end{align}
where in the last step we have used that the Gram determinant of the normal space generating mapping is $1$ such that $\pi_v(0) = q_{\rm{n}}(x|x)$. As $x$ was chosen arbitrarily on the manifold, this ends the proof.

\begin{lemma}\label{lemma:det_f} For $\tilde{f}$, $f$ and $x$ as defined above, we have that
	$\det G_{\tilde{f}}(x) = \det G_{{f}}(x)$.
\end{lemma}
\begin{proof}
	The Jacobian of $\tilde{f}$ is given by
	\begin{equation}\label{eq:jac_phi}
	J_{\tilde{f}}(u,v)= \left[ \begin{array}{c;{2pt/2pt}cc} J_f(u) + \frac{\partial}{\partial u}A_u v \ & \ A_u \end{array} \right] 
	%\left[ J_f(u) + \frac{\partial}{\partial u}A_u v \ ; \ A_u  \right] 
	\end{equation}
	where $\frac{\partial}{\partial u}$ denotes the Jacobian of a function depending on $u$, and the dashed line seperates two block matrices. Here we need that $f\in C^2$ to ensure the Jacobian is real. For points on the manifold is $v=0$, and thus the Gram determinant reduces to
	\begin{align}\label{eq:}
	\det G_{\tilde{f}}(x)&=\det \left( J_{\tilde{f}}(u,0)^T J_{\tilde{f}}(u,0) \right)   \notag \\
	&= \det
	\left[
	\begin{array}{c;{2pt/2pt}cc}
	J_f(u)^T J_f(u) & J_f(u)^T \cdot A_u\\ \hdashline[2pt/2pt]
	A_u^T \cdot J _f(u)^T   & A_u^T A_u
	\end{array}
	\right] \\
	&= \det
	\left[
	\begin{array}{c;{2pt/2pt}cc}
	J_f(u)^T J_f(u) & 0_{d\times {D-d}}\\ \hdashline[2pt/2pt]
	0_{{D-d}\times d}   & A_u^T A_u
	\end{array}
	\right]
	\\
	&= \det J_f(u)^T J_f(u) \cdot \det A_u^T A_u\\
	&= \det J_f(u)^T J_f(u) \\
	&= \det G_{f}(x)
	\end{align}
	where for the third equality we have exploited the fact that the column vectors of $J_f$ and $A_u$ are orthogonal. This was to be shown.	
\end{proof}

%\begin{remark}
%	Note that in Theorem \ref{thm:main}, we need that $\tilde{X}$ is diffeomorphic to $\mathbb{R}^D$. This requires that the noise distribution $q_{\rm{n}}(\cdot|x)$ is continuous for all $x$. 
%\end{remark}

\subsection{Proof of proposition \ref{prop:main}}
This follows immediately from the universality of standard NFs, see Section \ref{sec:problem}, and theorem \ref{thm:main}.

\subsection{Proof of statement in Remark \ref{rem:product_metric}}\label{ax:proof_product_metric}
We denote the probability measure of the random variable $X$ as  $\mathbb{P}_X$ and it is defined on $(\mathcal{X},\mathcal{B}(\mathcal{X}))$ where $\mathcal{B}(\mathcal{X})$ is the set of Borel sets in $\mathbb{R}^D$ intersected with $\mathcal{X}$. For a realisation of $X$, say $x$, we denote the probability measure of the shifted random variable $x+\mathcal{E}_{\rm{n}}$ as $\mathbb{P}_{\tilde{X}|X=x}$ and it is defined on $(\mathcal{N}_x,\mathcal{B}(N_x))$. We extend both measures to $(\mathbb{R}^D,\mathcal{B}(\mathbb{R}^D))$ by setting the probabilities to $0$ whenever a set $A\in\mathcal{B}(\mathbb{R}^D)$ has no intersection with $\mathcal{X}$ or $N_x$, respectively. 
For instance, that means for $\tilde{x}\in N_x$ that
\begin{align}\label{eq:}
\mathbb{P}[x+\mathcal{E}_{\rm{n}} \in (\tilde{x},\tilde{x}+d\tilde{x})] =&  \mathbb{P}[x+\mathcal{E}_{\rm{n}} \in (\tilde{x},\tilde{x}+d\tilde{x})\cap N_x] = \mathbb{P}_{\tilde{X}|X=x}[(\tilde{x},\tilde{x}+d\tilde{x})\cap N_x]
%
%\notag \\
% \notag\\
%=&\mathbb{P}[\varepsilon_{\rm{n}} \in (\epsilon,\epsilon+d\epsilon)] \notag \\
%=&\mathbb{P}_{\varepsilon_{\rm{n}}}[(\epsilon,\epsilon+d\epsilon)]
\end{align}
%$(\epsilon,\epsilon+d\epsilon)$ and
where  $(\tilde{x},\tilde{x}+d\tilde{x})$ denotes an infinitesimal volume element around $\tilde{x}$.
%, and $\mathbb{P}_{\varepsilon_{\rm{n}}}$ denotes the probability measure of $\varepsilon_{\rm{n}}$.

The mapping $(x,\varepsilon_{\rm{n}}) \mapsto x+\varepsilon_{\rm{n}}$ is $\mathcal{B}(\mathbb{R}^D) \times \mathcal{B}(\mathbb{R}^D)-$measurable, and thus
% because $\mathbb{R}^D$ is a topological vector space and $\mathcal{B}(\mathbb{R}^D)$ the Borel $\sigma-$algebra. What follows is that
$\tilde{X}=X+\mathcal{E}_{\rm{n}}$ is a random variable on $(\mathbb{R}^D,\mathcal{B}(\mathbb{R}^D))$  and has the pushforward of $\mathbb{P}_{(X,\mathcal{E}_{\rm{n}})}$ with regard to the mapping $(x,\varepsilon_{\rm{n}}) \to x + \varepsilon_{\rm{n}}$ as probability measure where $\mathbb{P}_{(X,\mathcal{E}_{\rm{n}})}$ is the joint measure of $X$ and $\mathcal{E}_{\rm{n}}$. Thus, for $A\in \mathcal{B}(\widetilde{\mathcal{X}})$, we have that
\begin{align}\label{eq:conv_dep_measures}
\mathbb{P}_{\tilde{X}}(A) &= \mathbb{P}_{(X,\mathcal{E}_{\rm{n}})}\left( \{(x,\varepsilon_{\rm{n}}) \in \mathbb{R}^D \times \mathbb{R}^D | x+\varepsilon_{\rm{n}} \in A \} \right). 
\end{align}
Now let $\tilde{x} \in N_x$ for an $x\in \mathcal{X}$.  Since $\mathcal{X}$ is $Q-$normally reachable,  $\mathbb{P}_{\tilde{X}}-$almost all $\tilde{x}$ are uniquely determined by $(x,\varepsilon_{\rm{n}})$ such that $\tilde{x} = x + \varepsilon_{\rm{n}}$. Therefore, we have for $\mathbb{P}_{\tilde{X}}-$almost all $\tilde{x}=x+\varepsilon_{\rm{n}}$  that
\begin{align}\label{eq:result_measures}
\mathbb{P}_{\tilde{X}}((\tilde{x},\tilde{x}+d\tilde{x})\cap \widetilde{\mathcal{X}})& =\mathbb{P}_{(X,\mathcal{E}_{\rm{n}})}\left( \{(x,\varepsilon_{\rm{n}}) \in \mathbb{R}^D \times \mathbb{R}^D | x+\varepsilon_{\rm{n}} \in (\tilde{x},\tilde{x}+d\tilde{x})\cap \widetilde{\mathcal{X}} \} \right)  \notag \\
& =\mathbb{P}\left( X+\mathcal{E}_{\rm{n}} \in (\tilde{x},\tilde{x}+d\tilde{x})\cap \widetilde{\mathcal{X}}  \right)  \notag \\
& = \mathbb{P} \left( X\in (x,x+dx) \cap \mathcal{X} \right)  \cdot \mathbb{P}\left( x+\mathcal{E}_{\rm{n}} \in (\tilde{x},\tilde{x}+d\tilde{x})\cap N_x  \right)  \notag \\
%& =\mathbb{P}_{(X,\varepsilon_{\rm{n}})}\left( \{(x,\epsilon) \in \mathbb{R}^D \times \mathbb{R}^D | x+\epsilon \in (\tilde{x},\tilde{x}+d\tilde{x}) \cap N_x  \} \right) \\
&=  \mathbb{P}_{X} \left(  (x,x+dx) \cap \mathcal{X} \right)  \cdot \mathbb{P}_{\tilde{X}|X=x}\left(  (\tilde{x},\tilde{x} + d\tilde{x})\cap N_x\right)
\end{align}
where for the first equality we used equation (\ref{eq:conv_dep_measures}) and for the third the fact that $(x,\varepsilon_{\rm{n}})$ is almost surely uniquely determined by $\tilde{x}$.

Both probability measures on the right-hand side have a density. For $\mathbb{P}_X$ with respect to $dV_f$, see Section \ref{sec:problem}, this density is $p^*(x)$. Similarly, since $N_x$ is a linear subspace of $\mathbb{R}^D$,  $q_{\rm{n}}(\tilde{x}|x)$ is the density of $\mathbb{P}_{\tilde{X}|X=x}$ with respect to a volume form $dV_{h}$ where $h$ is the mapping from $\mathbb{R}^{D-d}$ to  $N_x$. Then, the corresponding density of $\mathbb{P}_{\tilde{X}}$ with respect to the product measure $V_{\otimes}:=V_f\otimes V_h$ is given by
\begin{align}\label{eq:result_dens}
q^{\otimes}_{\rm{n}}(\tilde{x})& = p^*(x) q_{\rm{n}}(\tilde{x}|x)%p(h^{-1}(y)) %q_{\rm{n}}(\epsilon|x),
\end{align}
and it holds that
\begin{align}\label{eq:}
\int_{\widetilde{\mathcal{X}}} q^{\otimes}_{\rm{n}}(\tilde{x}) dV_{\otimes}(\tilde{x})=& \int_{\mathcal{X}} \int_{N_x}p^*(x) q_{\rm{n}}(\tilde{x}|x)  dV_h(\tilde{x}) dV_f(x) \notag \\
=& \int_{\mathcal{X}} p^*(x) dV_f(x) \notag \\
=& 1, 
\end{align}
as needed for a density on $\widetilde{\mathcal{X}}$. This ends the proof.

\subsection{Proof of proposition \ref{prop:nois_choice_JP}}\label{ax:proof_prop}
The generating function $f$ is an embedding for $\mathcal{X}$ and $X=f(u)$ has the density $p^*(x)$ for $x\in \mathcal{X}$. We may extend the domain of $p^*(x)$ to include all points $x\in \mathbb{R}^D$ using the Dirac-delta function. We denote this density with $\bar{p}(x)$ at it is given by 
\begin{equation}\label{eq:before_eq}
\bar{p}(x) = \int_{\mathcal{U}}  \delta(x-f(u))\pi_u(u) dz,
\end{equation}
see \cite{general_dirac}.
After inflating $X$, we have that 
\begin{equation}\label{eq:after_eq}
p_{\Sigma} (\tilde{x})= \int_{\mathcal{U}} \mathcal{N}(\tilde{x};f(u), \Sigma) \pi_u(u) dz
\end{equation}
with covariance matrix $\Sigma \in \mathbb{R}^{D\times D}$ where for $\Sigma = \sigma^2 I$ we have that $\lim_{\sigma \to 0} p_{\Sigma} (\tilde{x}) = \bar{p}(\tilde{x})$. Assume $\tilde{x}=x$ for some $x\in \mathcal{X}$. We Taylor expand $f(u)$ around $u_0=f^{-1}(x)$ up to first order, 
\begin{equation}\label{eq:}
f(u) \approx f(u_0) + J_f(u_0) (u-u_0),
\end{equation}
and $\pi_u(u)$ up to second order,
\begin{equation}\label{eq:}
\pi_u(u) \approx \pi_u(u_0) + \pi_u(u_0)' (u-u_0) + \frac12 (u-u_0)^T \pi_u ''(u_0) (u-u_0).
\end{equation}
where $\pi_u(u_0)'$ denotes the gradient  and $\pi_u''(u_0)$ the Hessian of $\pi$ evaluated at $u_0$, thus $\pi_u(u_0)'\in \mathbb{R}^d$ and $\pi_u''(u_0)\in \mathbb{R}^{d\times d}$. Then, we can approximate $p_{\Sigma} (x)$ as follows:
\begin{align}\label{eq:}
p_{\Sigma} (x)  \approx& \frac{1}{\sqrt{(2\pi)^D \det(\Sigma)}} \int_{\mathcal{U}} \exp\left(-\frac12 (u-u_0)^T J_f ^T \Sigma^{-1}J_f(u-u_0) \right) \cdot \notag \\
&\cdot \left(\pi_u(u_0) + \pi_u'(u_0)^T (u-u_0) + \frac12 (u-u_0)^T \pi_u ''(u_0) (u-u_0) \right) dz.
\end{align}
Now define $\hat{\Sigma}^{-1} = J_f^T \Sigma^{-1} J_f$. Then,
\begin{align}\label{eq:}
p_{\Sigma} (x)  \approx& \frac{\sqrt{ \det(\hat{\Sigma})}}{\sqrt{ (2\pi)^{D-d} \det(\Sigma)}} \int_{\mathcal{U}} \frac{1}{\sqrt{(2\pi)^d \det(\hat{\Sigma})}} \exp\left(-\frac12 (u-u_0)^T  \hat{\Sigma}^{-1}(u-u_0) \right)\cdot \notag \\
&\cdot(\pi_u(u_0) + \pi_u'(u_0)^T (u-u_0) + \frac12 (u-u_0)^T \pi_u ''(u_0) (u-u_0)) dz.
\end{align}
Thus, we can exploit the Gaussian in $\mathcal{U}$-space and get
\begin{align}\label{eq:}
p_{\Sigma} (x) \approx&  \frac{\sqrt{ \det(\hat{\Sigma})}}{\sqrt{ (2\pi)^{D-d} \det(\Sigma)}}  (\pi_u(u_0) + \frac12 \mathbb{E}\left[ (u-u_0)^T \pi_u''(u_0) (u-u_0) \right] ) \notag \\
=& \frac{\sqrt{ \det(\hat{\Sigma})}}{\sqrt{ (2\pi)^{D-d} \det(\Sigma)}}  (\pi_u(u_0) + \frac{1}{2} ||\pi_u''(u_0) \odot \hat{\Sigma}||_+),
\end{align}
where $\odot$ stands for the elementwise multiplication and $||A||_+=\sum_{i,j=1}^{d} A_{ij}$ for a $\mathbb{R}^d\times \mathbb{R}^d$ matrix $A$.

For the special case where $\Sigma = \sigma^2 I_D$, we can simplify this expression by exploiting that

\begin{align}\label{eq:}
\frac{\sqrt{ \det(\hat{\Sigma})}}{\sqrt{(2\pi)^{D-d} \det(\Sigma)}}& = \frac{1}{(2\pi)^{\frac{D-d}{2}}} \frac{\sigma^{-D}}{\sigma^{-d} \sqrt{\det G_f}} \notag \\
&= \frac{1}{(2\pi \sigma^2)^{\frac{D-d}{2}} \sqrt{\det G_f}}.
\end{align}
Thus, in total, we get for this special choice of $\Sigma$

\begin{align}\label{eq:taylor_end}
p_{\sigma} (x)  \approx&  \frac{1}{(2\pi \sigma^2)^{\frac{D-d}{2}} \sqrt{\det G_f}}(\pi_u(u_0) + \frac{\sigma^2 }{2} ||\pi_u''(u_0) \odot (J_f^T J_f)^{-1}||_+) \notag \\
=&\frac{1}{(2\pi \sigma^2)^{\frac{D-d}{2}} \sqrt{\det G_f}} \pi_u(u_0)( 1 + \frac{\sigma^2}{2 \pi_u(u_0)} ||\pi_u''(u_0) \odot (J_f^T J_f)^{-1}||_+)
\end{align}
%We want to show that $\lim_{\sigma \to 0} p_\sigma(x) = p^*(x)$. For that 
We assume now
\begin{equation}\label{eq:}
\left| \frac{\sigma^2}{2 \pi_u(u_0)} ||\pi_u''(u_0) \odot (J_f^T J_f)^{-1}||_+ \right| \ll 1.
\end{equation}
%
%Ignoring the prefactor $1 /  (2\pi \sigma^2)^{\frac{D-d}{2}}$ for a moment, the first term on the right-hand side corresponds to the change of variable formula. Thus, we want the second term is small. This was to be shown. 
Note that $1/(2\pi \sigma^2)^{\frac{D-d}{2}}$ from equation (\ref{eq:taylor_end}) is exactly the normalization constant obtained when inflating the manifold with Gaussian noise in the normal space, $q_{\rm{n}}(x|x) = 1/(2\pi \sigma^2)^{\frac{D-d}{2}}$. What follows is that $\lim_{\sigma \to 0}p_{\sigma}(x)/q_{\rm{n}}(x|x) = p^*(x)$ as we wanted to show.

\subsection{Proof of theorem \ref{thm:reach}} \label{ax:proof_reach}
The result follows directly from the definition of the reach number $\tau_{\mathcal{X}}$ of $\mathcal{X}$. It is defined as the supremum of all $r\geq 0$ such that the orthogonal projection $\text{pr}_{\mathcal{X}}$ on $\mathcal{X}$ is well-defined on the $r-$neighbourhood $\mathcal{X}^{r}$ of $\mathcal{X}$,
\begin{equation}\label{eq:}
\mathcal{X}^{r}:=\{\tilde{x}\in \mathbb{R}^D | \ \text{dist}(\tilde{x},\mathcal{X})\leq r  \}
\end{equation}
where $\text{dist}(\tilde{x},\mathcal{X})$ denotes the distance of $\tilde{x}$ to $\mathcal{X}$.
Thus,
\begin{equation}\label{eq:}
\tau_{\mathcal{X}} = \sup \left\lbrace r\geq 0 \ | \ \forall \tilde{x}\in \mathbb{R}^D, \ \text{dist}(\tilde{x},\mathcal{X}) \leq r \implies \exists! x\in \mathcal{X} \text{ s.t. dist}(\tilde{x},\mathcal{X})=||\tilde{x}-x||  \right\rbrace ,
\end{equation}
see Definition 2.1. in \citet{reach_density}. By assumption $\tau_{\mathcal{X}}>0$. Thus for all $\tilde{x}\in \mathcal{X}^{\tau_{\mathcal{X}}}$ we have that $x:=\text{pr}_{\mathcal{X}}(\tilde{x})$ is unique. Since $\mathcal{X}$ is a closed manifold, it must hold that $\tilde{x}\in N_x$ where $N_x$ denotes the normal space in $x$. Let the noise generating distributions be a uniform distribution on the ball with radius $\tau_{\mathcal{X}}$, thus 
\begin{equation}\label{eq:}
q_{\rm{n}}(\tilde{x}|x)= \text{Uniform}(\tilde{x}; B(x,\tau_{\mathcal{X}})\cap N_x),
\end{equation}
where $B(x,\tau_{\mathcal{X}})$ denotes  a $D-$dimensional ball with radius $\tau_{\mathcal{X}}$ and center $x$. Then, we have for $\widetilde{\mathcal{X}}=\bigcup_{x\in \mathcal{X}} N_{q_{\rm{n}}(\cdot |x)}$ that
\begin{equation}\label{eq:}
\widetilde{\mathcal{X}} = \mathcal{X}^{\tau_{\mathcal{X}}}.
\end{equation}
Thus, $\mathcal{X}$ is $Q-$normally reachable where $Q:=\{q_{\rm{n}}(\cdot|x)\}_{x\in\mathcal{X}}$.

%% file: JMLR_appendix2.tex
\section{Experiments}\label{AX:experiments}
For all expriments, we use Adam optimizer with an initial learning rate 0.1, a learning rate decay of 0.5 after 2000 optimization steps without improvement (learning rate patience). We use a batch size of $200$. No hyperparameter fine-tuning was done.
\subsection{Technical details for circle experiments} \label{AX:experiment_1}
We use  a BNAF (Block Neural Autoregressive Flow) to learn the inflated density, see table \ref{tbl:2D_von_Mise} for the details. There, we report the number of hidden layers, hidden dimensions (which scales with the dimensionality of the embedding space), total parameters of the model, and, finally, the number of gradient steps (iterations).

For the FOM and $\chi^2-$noise models, we use the same architecture as for the $D=2$ case.
\begin{center}
	\begin{table}[H]
		\begin{tabular}{|c c c c c|}
			\hline
			Data dimension & hidden layers & hidden dimension & total parameters & iterations \\ 
			\hline\hline
			2 & 3 & 100 & 31,204 & 70000\\ 
			\hline
			5 & 3 & 250 &192,010 & 70000  \\
			\hline
			10 & 3 & 500 &764,000 & 70000  \\
			\hline
			15 & 3 & 750 & 1,716,030  & 100000  \\
			\hline
			20 & 3 & 1000 &3,048,040 & 100000  \\
			\hline
		\end{tabular}
		\caption{BNAF details for circle experiments.}
		\label{tbl:2D_von_Mise}
	\end{table}
\end{center}

\subsection{Technical details for density estimation tasks} \label{ax:de_on_manifolds}
We use  a BNAF (Block Neural Autoregressive Flow) to learn the inflated density, see table \ref{tbl:BNAF_surface} for the details. There, we report the number of hidden layers, hidden dimensions, total parameters of the model, and, finally, the number of gradient steps (iterations). 
\begin{center}
	\begin{table}[H]
		\begin{tabular}{|c c c c c|}
			\hline
			Data dimension & hidden layers & hidden dimension & total parameters & iterations \\ 
			\hline\hline
			1 & 6 & 210 & 31,204 & 50000\\ 
			\hline
			2 & 6 & 210 &268,384 & 50000  \\
			\hline
			3 & 6 & 210 &268,806 & 50000  \\
			\hline
			4 & 6 & 200 &244,408 & 50000  \\
			\hline
		\end{tabular}
		\caption{BNAF details for density estimation experiments.}
		\label{tbl:BNAF_surface}
	\end{table}
\end{center}

\subsubsection{Latent densities}\label{ax:latent_densities}
In table \ref{tbl:latent_densities} we show the latent densities used in the experiments in order of appereance.

\begin{table}[H]
	\centering
	\resizebox{\columnwidth}{!}{
		\begin{tabular}{lll}
			\toprule
			%		\multicolumn{5}{c}{StyleGAN $d=2$ \qquad \qquad \qquad  \qquad \qquad  StyleGAN $d=64$}       \\ 
			%		\cmidrule(r){1-7}
			Manifold & $\pi_u(u) \propto $ & Parameters \\
			\midrule
			$\mathbb{S}^2$ & $\sum_{i=1}^{4} \exp (6 \cos(u_1 - \mu_i)) \cdot  \exp(6\cos(2(u_2-m_i)))$ & table \ref{tbl:S2_mixture}  \\
			\cdashline{2-3}
				 & $\sum_{i=1}^{2} \exp (6 \cos(u_1 - \mu_i)) \cdot  \exp(6\cos(2(u_2-m_i))) + \frac{1}{2\pi} \cdot 2\exp(50\cos(2(u_2-m_3)))$ & table \ref{tbl:S2_correlated}\\
			\midrule
			$\mathbb{T}^2$ & 		$\sum_{i=1}^{3} \exp (2 \cos(u_1 - \mu_i)) \cdot  \exp (2 \cos(u_2 - m_i))$ & table \ref{tbl:T2_mixture}  \\
			\cdashline{2-3}
		    & 		$\frac{1}{2\pi}\exp (2 \cos(u_1+u_2 - 1.94))$ &   \\
			\midrule
			$\mathbb{H}^2$ & 		 $2\exp\left(-\frac{u_1}{2}\right) \frac{1}{2\pi}$ &  \\
			\cdashline{2-3}
						  & 		 $\frac{1}{2}\exp(6 \cos(u_2-u_1-\pi))$ &  \\
			\midrule
			thin spiral & $\frac{1}{0.3} \exp(0.3z)$ &   \\
		\midrule
			swiss roll & $\sum_{i=1}^{3} \exp (\kappa \cos(2\pi u_1 - \mu_i)) \cdot  \exp (\kappa \cos( 2\pi u_2 - \mu_i))$ & table \ref{tbl:swiss_mixture} \\
			\cdashline{2-3}
			& $ \frac{1}{2\pi}\cdot 2\pi \exp (\kappa \cos(2\pi (u_2 - u_1)))$ &   \\
			\midrule
			$\mathbb{(HS)}^2$ & $\sum_{i=1}^{3} \exp (\kappa \cos(u_1 - \mu_i)) \cdot  \exp (\kappa \cos( u_2 - \mu_i))$ & \\	
			\cdashline{2-3}
			& $\frac{1}{2} \exp(0.3|u_1|) \exp(\kappa \cos(u_2-u_1-\pi))$ & \\	
			\midrule
			$SO(2)$ & $\sum_{i=1}^{4} \exp (6 \cos(u_1 - \mu_i)) \cdot  \exp(6\cos(u_2-m_i))$ & table \ref{tbl:stiefel} 	\\
			\bottomrule 
		\end{tabular}
	}
	\caption{Latent densities.}
	\label{tbl:latent_densities}
\end{table}

\begin{table}[H]
	\centering	
	\begin{tabular}{|c c c|}
		\hline
		i & $\mu_i$ & $m_i$ \\ 
		\hline\hline
		1 & $\frac{\pi}{2}$ & $\frac{\pi}{4}$\\ 
		2 & $\frac{\pi}{2}$ & $\frac{3\pi}{4}$\\ 
		3 & $\frac{3\pi}{2}$ & $\frac{\pi}{4}$\\  
		4 & $\frac{3\pi}{2}$ & $\frac{3\pi}{3}$\\  
		\hline
	\end{tabular}
	\caption{Mixture parameters of von Mises on $\mathbb{S}^2$.}
	\label{tbl:S2_mixture}
\end{table}

\begin{table}[H]
	\centering	
	\begin{tabular}{|c c c|}
		\hline
		i & $\mu_i$ & $m_i$ \\ 
		\hline\hline
		1 & $0$ & $\frac{\pi}{2}$\\ 
		2 & $\pi$ & $\frac{3\pi}{2}$\\ 
		\hline
	\end{tabular}
	\caption{Mixture parameters of von Mises on $\mathbb{S}^2$.}
	\label{tbl:S2_correlated}
\end{table}

\begin{table}[H]
	\centering	
	\begin{tabular}{|c c c|}
		\hline
		i & $\mu_i$ & $m_i$ \\ 
		\hline\hline
		1 & $0.21$ & $2.85$\\ 
		2 & $1.89$ & $6.18$\\ 
		3 & $3.77$ & $1.56$\\
		\hline
	\end{tabular}
	\caption{Mixture parameters of von Mises on $\mathbb{T}^2$.}
	\label{tbl:T2_mixture}
\end{table}

\begin{table}[H]
	\centering	
	\begin{tabular}{|c c c|}
		\hline
		i & $\mu_i$ & $m_i$ \\ 
		\hline\hline
		1 & $0.1$ & $0.1$\\ 
		2 & $0.5$ & $0.8$\\ 
		3 & $0.8$ & $0.8$\\
		\hline
	\end{tabular}
	\caption{Mixture parameters of von Mises on swiss roll.}
	\label{tbl:swiss_mixture}
\end{table}

\begin{table}[H]
	\centering	
	\begin{tabular}{|c c |}
		\hline
		i & $\mu_i$  \\ 
		\hline\hline
		1 & $0$\\ 
		2 & $-\frac{\pi}{2}$\\ 
		3 & $\frac{\pi}{2}$ \\  
		4 & $\pi$ \\  
		\hline
	\end{tabular}
	\caption{Mixture parameters of von Mises on $SO(2)$.}
	\label{tbl:stiefel}
\end{table}

\subsection{Density estimation on MNIST digit 1}\label{ax:de_on_MNIST}
%We used the original implementation of the $\mathcal{M}-$flow. 
For fair comparison, we tried to use the same architectures for the IID and $\mathcal{M}-$flow. As the latter requires to invert the flow during training, we have used rational-quadratic splines\footnote{An interval $[-B,B]$ is split into $K$ equidistant bins, and on each subinterval, a rational-quadratic spline is defined such that the derivatives are continuous at the boundary points. The parameters of the splines are again outcomes of neural networks. We refer to $B$ as the spline range and $K$ as the bin size in the following. Outside of the interval $[-B,B]$, the transformation is set to the identity.} for the flows which can be efficiently inverted, see \citet{durkan2019neural} and  table \ref{tbl:MNIST_architectures}. Note that the $\mathcal{M}-$flow learns the latent density using an additional flow $f_{\phi}$ after learning to reconstruct the data using $f_{\psi}$, see \citet{brehmer2020flows}.

\begin{table}[h]
	\centering
	\resizebox{\columnwidth}{!}{
		\begin{tabular}{lllllll}
			\toprule
			\multicolumn{4}{c}{flow $f_{\psi}$} & {flow $h_{\phi}$} \\
			%		\multicolumn{8}{c}{flow $h_{\phi}$} 
			\cmidrule(r){1-3}
			\cmidrule(r){4-5}
			model        & \# couplings & coupling type &     \# couplings  & coupling type & \#paramters  \\
			\midrule
			IID  & 10  & spline with $B=11$,$K=10$  &  - & - & 14M    \\
			$\mathcal{M}$-flow   & 10  & spline with $B=11$,$K=10$  &  10  & spline with $B=11$,$K=10$  & 14.3M     \\
			\bottomrule \\
		\end{tabular}
	}
	\caption{Architectures for standard IID and $\mathcal{M}-$flow on MNIST digit 1.}
	\label{tbl:MNIST_architectures}
\end{table}

We train on 100 epochs with a batch size of 100, and take the model yielding the best result on the validation set (10\% of training set). We use AdamW optimizer (\cite{loshchilov2017decoupled}) and anneal the learning rate to $0$ after 100 epochs using a cosine schedule (\cite{loshchilov2016sgdr}). 

We apply weight decay with a prefactor of $10^{-6}$ without dropout. Furthermore, a $L_2-$regularization on the latent variable with a prefactor of $0.01$ was used to stabilize the training. 

\subsubsection{Additional Figures} \label{ax:additional_figure}

\begin{figure}[H]
	\centering %  width=1\linewidth,height=1\pageheight
	\vspace*{-1cm}
	\includegraphics[width=\textwidth]{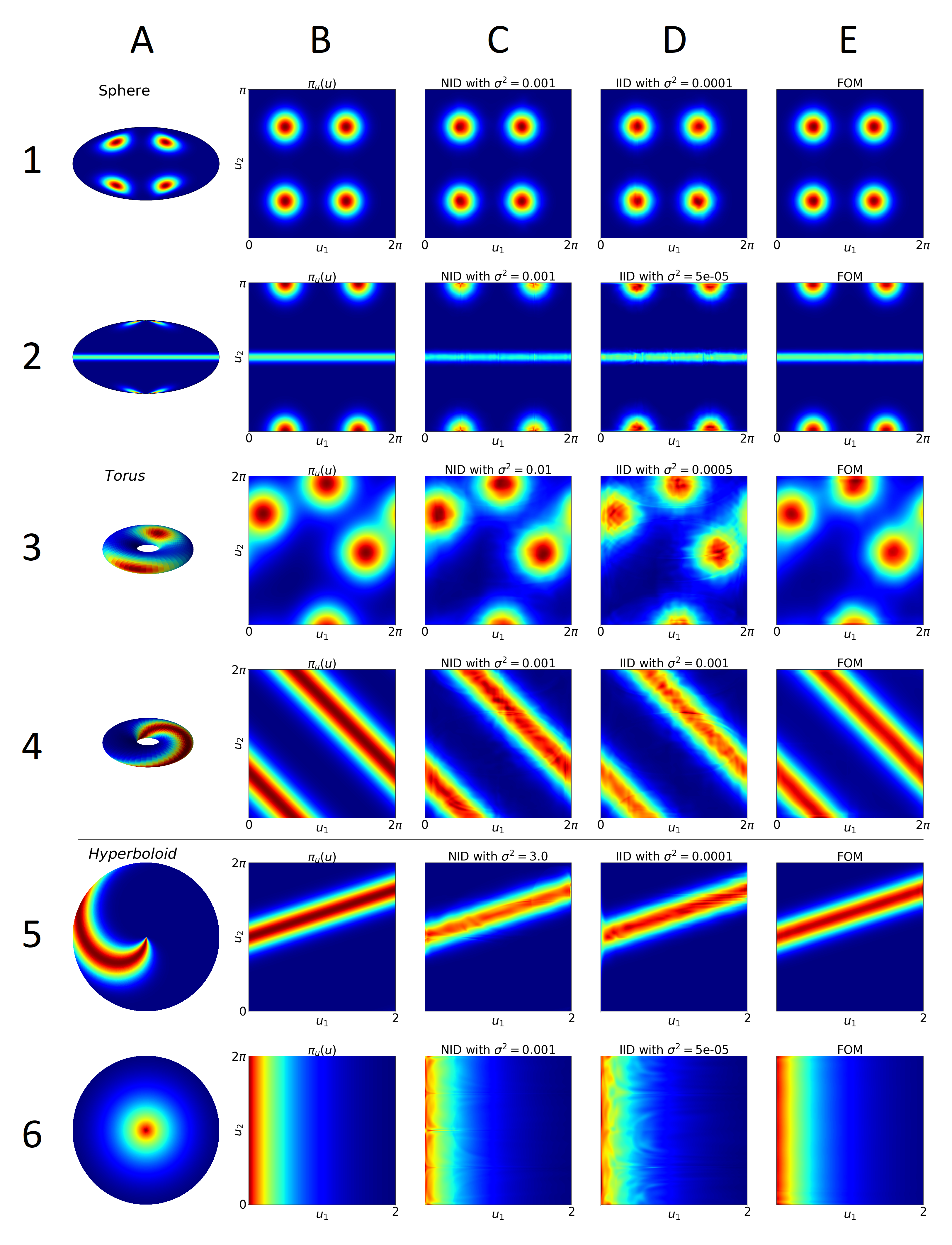}
%	\vspace*{-2.5cm}
	\caption{\textbf{Columns A and B:} Target density in data (\textsl{A}) and latent space (\textsl{B}) for various manifolds and different latent distributions. \textbf{Column C: }Best learned density using our method with NID. \textbf{Column D: }Best learned density using our method with IID.  \textbf{Column D: }Learned density using FOM.}
%	\caption{\textbf{First two columns:} Target density in data (\textsl{first}) and latent space (\textsl{second}) for various manifolds and different latent distributions. For $p^*(x)$ of the sphere, we use the Mollweide project. For the Hyperboloid, we project onto the Poincaré ball.  \textbf{Last two columns: }Best learned density using our method with NS (\textsl{third}). KS vs. $\sigma^2$ plot for FG and NS (we used 3 seeds for the error bars)  with the KS value of FOM as horizontal line (\textsl{last}). \textbf{table: }Optimal KS values for the different models.}
	\label{fig:DEmanifolds_appendix0}
\end{figure}

\begin{figure}[H]
	\centering %  width=1\linewidth,height=1\pageheight
	\vspace*{-1cm}
	\includegraphics[width=\textwidth]{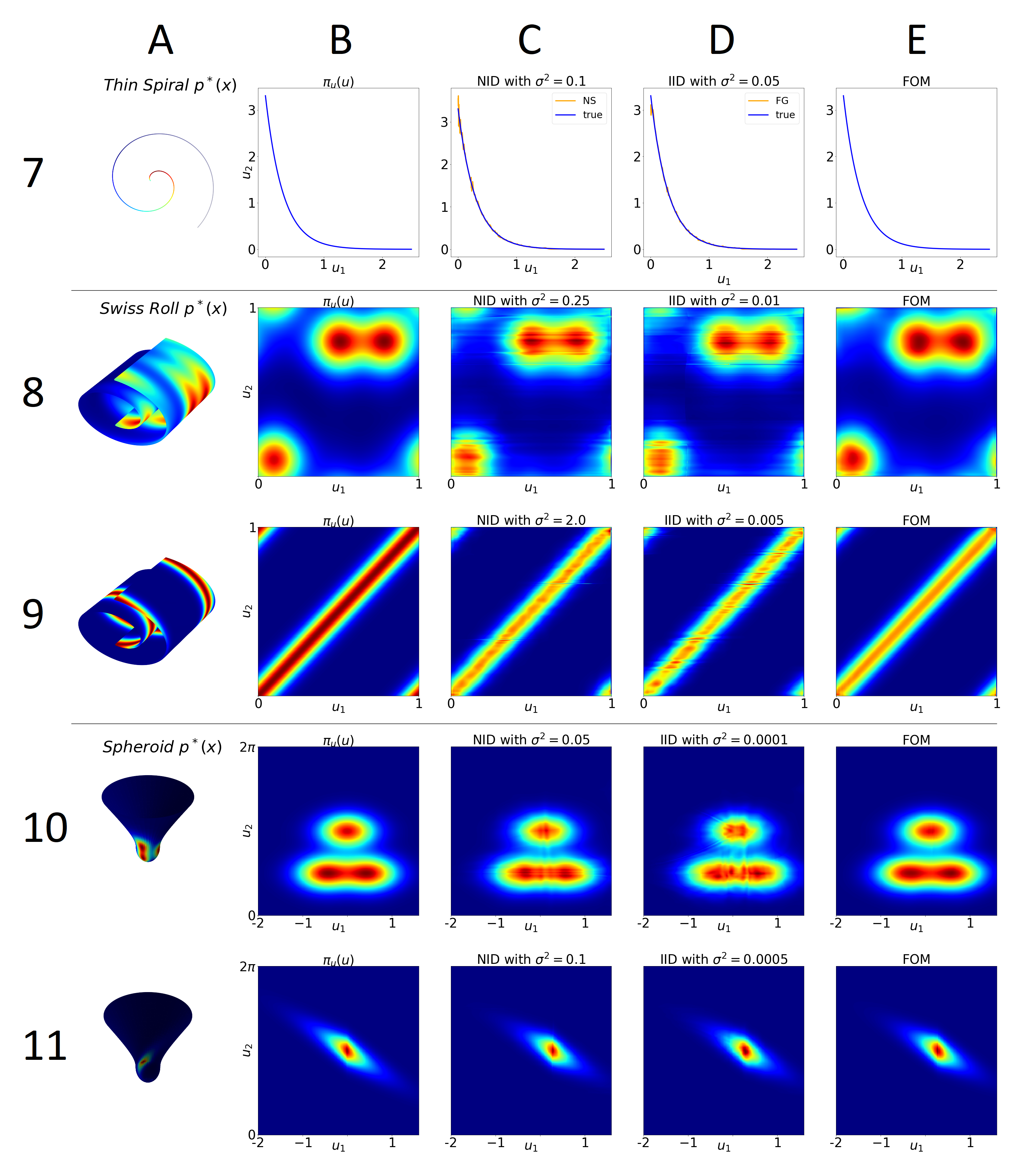}
%	\vspace*{-2cm}
	\caption{\textbf{Columns A and B:} Target density in data (\textsl{A}) and latent space (\textsl{B}) for various manifolds and different latent distributions. \textbf{Column C: }Best learned density using our method with NID. \textbf{Column D: }Best learned density using our method with IID.  \textbf{Column D: }Learned density using FOM.}
	\label{fig:DEmanifolds_appendix1}
\end{figure}

\begin{figure}[H]
	\centering
	\includegraphics[width=\textwidth]{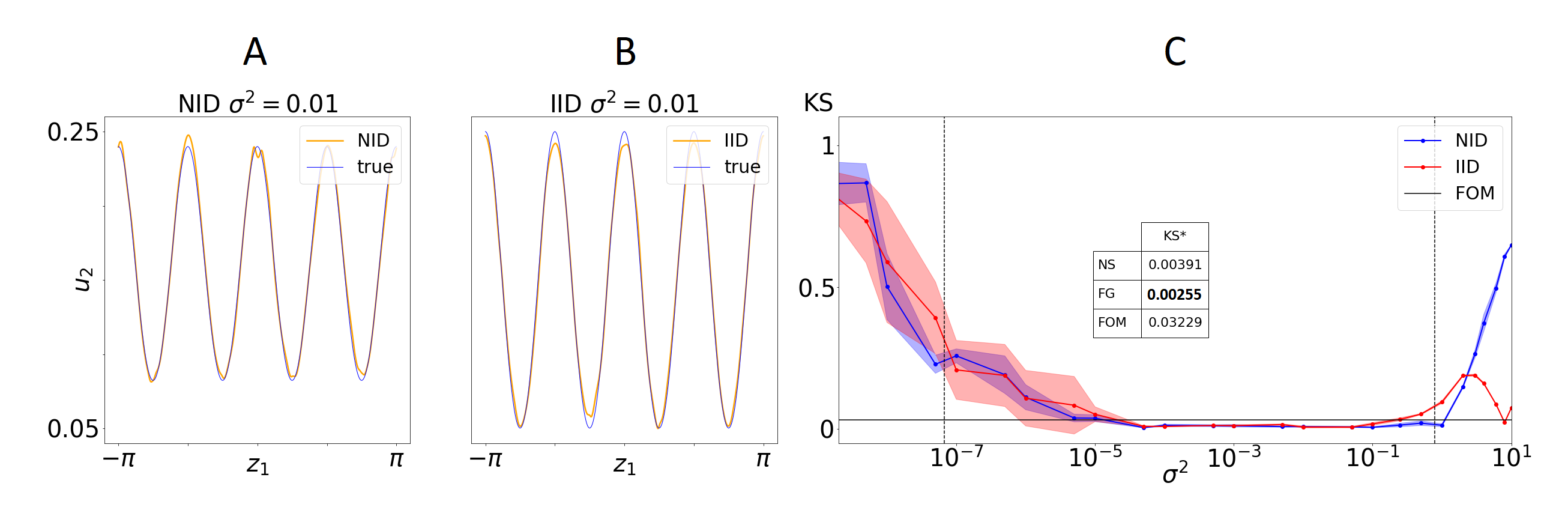}
	\caption{\textbf{A: } Learned latent density using NID (\textsl{orange}) on top of target latent density (\textsl{blue}). \textbf{B: }Learned latent density using IID (\textsl{orange}) on top of target latent density (\textsl{blue}).   \textbf{C: } KS vs. $\sigma^2$ plot for IID and NID (we used 3 seeds for the error bars)  with the KS value of FOM as horizontal line. \textbf{Table in C: }Optimal KS values for the different models. \textbf{Vertical Lines in C: }Lower and upper bound (see point \ref{standard_procedure_sigma_bounds} of the standard procedure).}
	\label{fig:DEstiefel001}
\end{figure}

We show the performance of the IID, FOM together with the NID method on the density estimation tasks from Section \ref{sec:DE_manifolds} in Figure \ref{fig:DEmanifolds_appendix0} and \ref{fig:DEmanifolds_appendix1}. In Figure \ref{fig:DEstiefel001}, we show the learned latent density using NS on the $SO(2)$ with $\sigma^2=0.01$, as motivated in the main text. 

%Finally, in Figure \ref{fig:MNIST_dequatization_flip}, we show the MNIST log-densities using different dequatization schemes. Different to Figure \ref{fig:MNIST}, zero-mean Gaussian has been added to the training images before applying the $\mathcal{M}-$flow, whereas no dequatization has been performed for the IID. Again the areas of intersection are smilar, for IID roughly 0.064, and for the M-flow roughly 0.07 .We conclude that this dequatization noise does not change the results substantially.

 \begin{figure}[h]
	\centering					
	\includegraphics[scale=0.25]{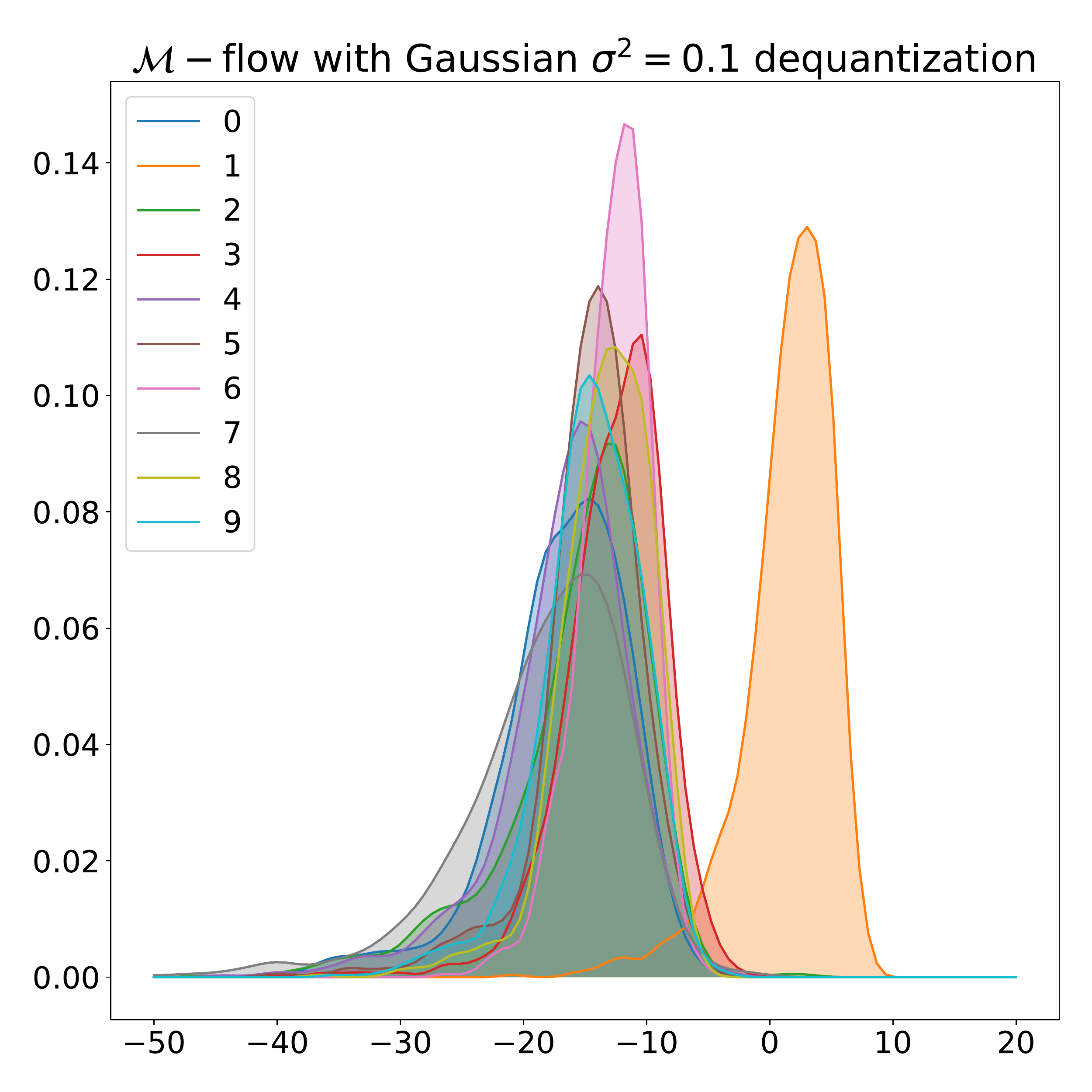}
	\includegraphics[scale=0.25]{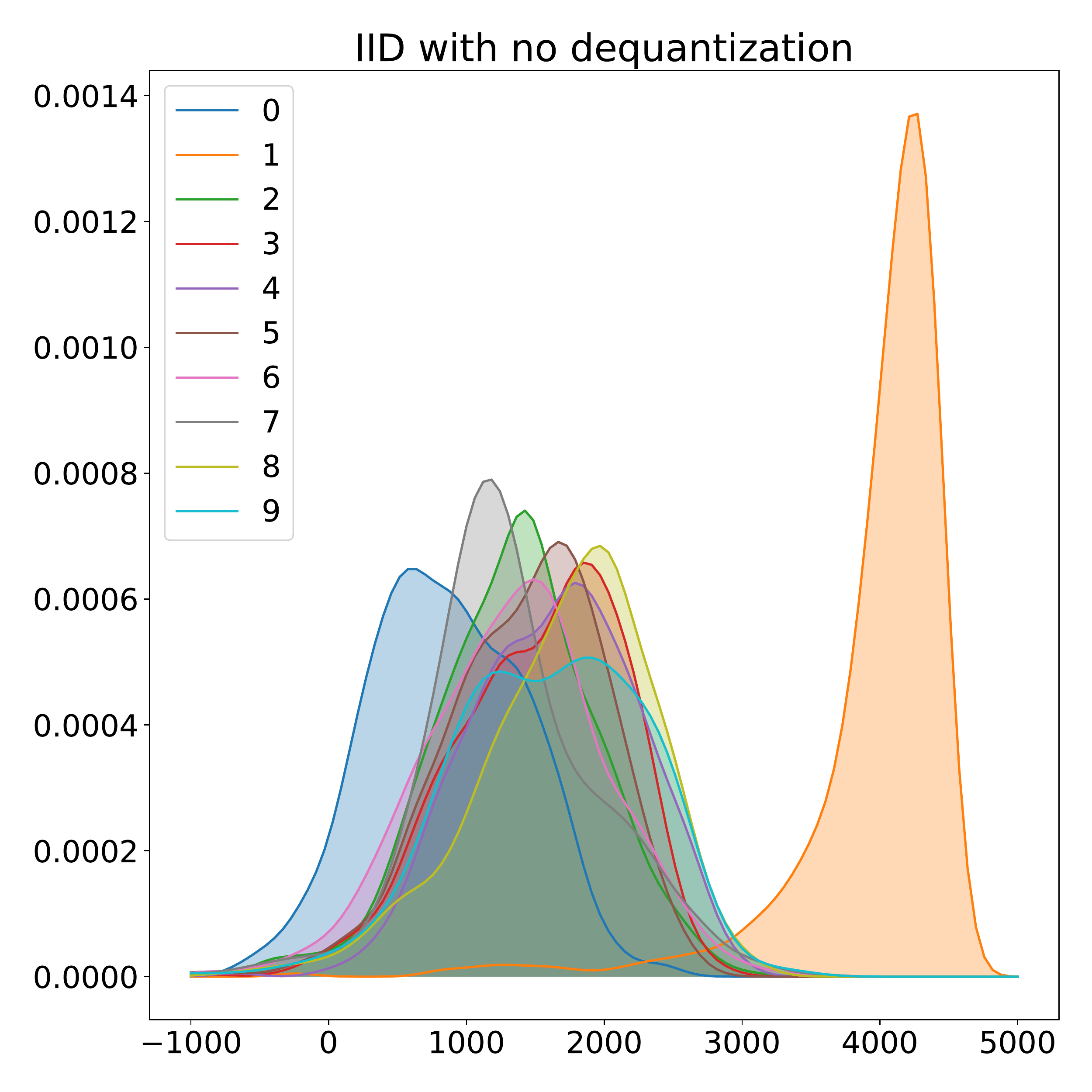}
	\caption{Log likelihoods on various MNIST test digits using $\mathcal{M}-$flow (\textbf{left}) and IID (\textbf{right}) trained on digit 1 only. }
	\label{fig:MNIST_dequatization_flip}
\end{figure}

\subsection{Manifold Flow for the mixture of von Mises distributions on $S^2$} \label{AX:sphere_related_work}
%\subsubsection{Related work on $S^2$}\label{subsec:related_work_performance}
In this Subsection, we apply the manifold flow, see Section \ref{sec:related_work}, on a mixture of von Mises distributions on a sphere. We do not attempt to find the optimal hyperparameters and training settings (such as batch- and training size, optimization method, or training scheduler) to maximize the performance.

\begin{figure}[h]
	\centering
	\includegraphics[scale=0.23]{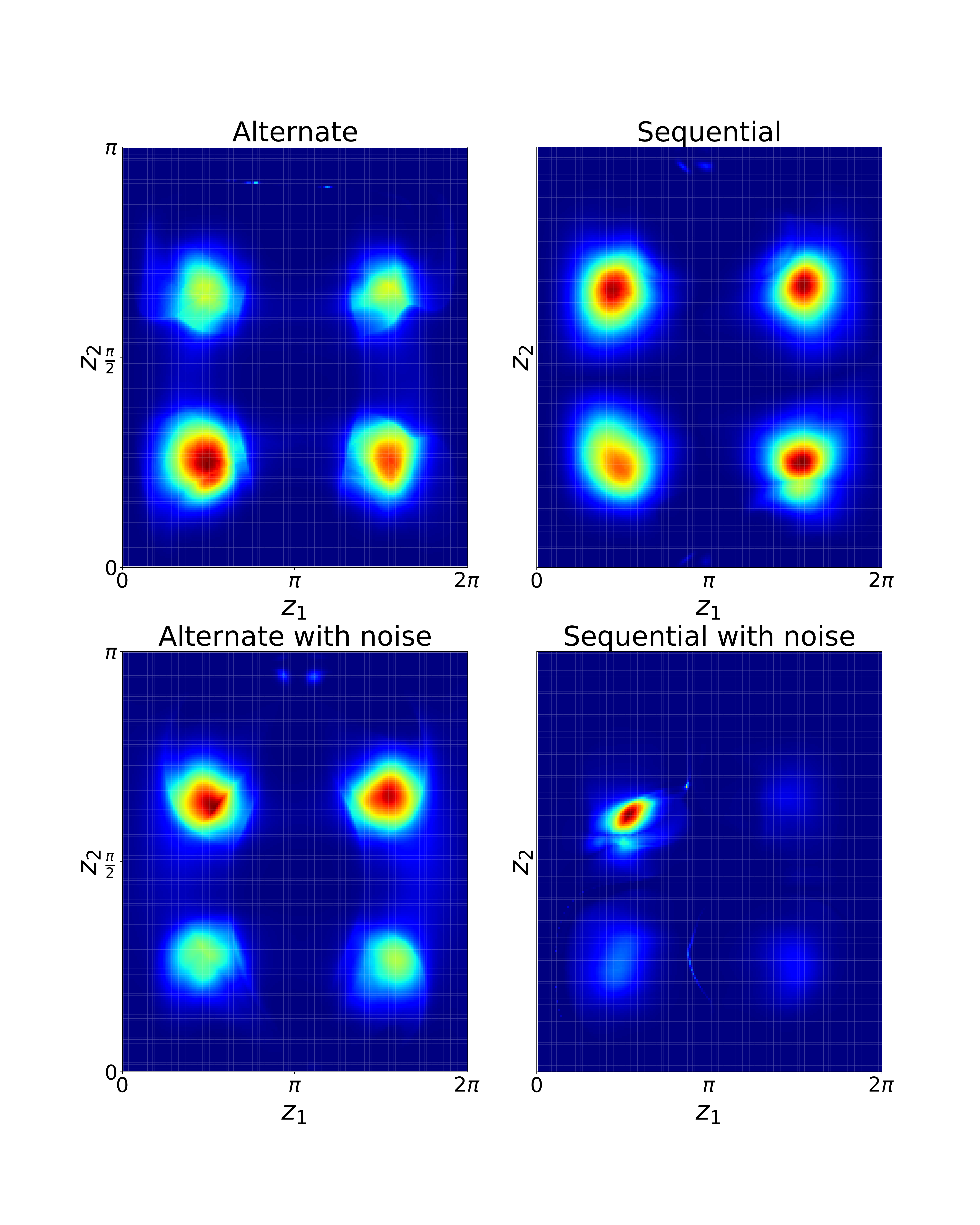}
	\caption{Performance of MF on the mixture of von Mises distributions on a sphere (top) and noisy sphere (bottom), using different training schemes (alternating left, and sequential right).}
	\label{MF_no_noise}
\end{figure}

%Our aim is to clarify the main differences to our approach on a conceptual basis. 
% We restrict ourselves to the manifold flow and PIE model, which are the closest to our approach.
%Note that the actual performance (as e.g. measured using the KS statistics) depend on many hyperparameters .
% Since all of the proposed methods depend on state of the art Neuronal Network architectures, for which the performance depend on many hyperparameters (such as the batch size, the optimization method or training scheduler), we do aim to compare performances, but rather conceptual differences.
%Finally, we show how related work, see Section \ref{sec:related_work}, perform on this mixture of von Mises on a sphere. For more training details, such as model architectures, we refer to Appendix \ref{AX:sphere_related_work}.
The manifold flow (MF) proposed by \Citealp{brehmer2020flows} uses two flows, one for encoding the data manifold to the latent space, and another for learning the latent density. To avoid calculating the Gram determinant of the encoding flow, they proposed different training procedures, an alternating, and a sequential (see Section \ref{sec:related_work} for more details). In Figure \ref{MF_no_noise}, we show that both methods learn the density reasonable good (top left and right). However, if we add Gaussian noise with magnitude $0.01$ to the dataset, the two training schemes lead to very different results (bottom left and right). This illustrates the drawback of not having a unified maximum likelihood objective.  We used the same model and training settings as for a similar dataset (a two-dimensional manifold embedded in $\mathbb{R}^3$) studied in \Citealp{brehmer2020flows}.
%performs worse compared to FOM or our method, see Figure \ref{MF_on_sphere}. Note that \Citealp{brehmer2020flows} trains the MF on a fixed training set, whereas we sample for each gradient update a new set of samples. Thus, for fair comparison, we also train the MF using a sampling based approach, which significantly improves the performance, see Appendix \ref{AX:sphere_related_work} for the original results. We have found that the sequential training method (see Section \ref{sec:related_work} for details) did not even converge using a training set of $10^5$ samples, which illustrates the drawback of not having a unified maxixum likelihood objective. 
%, and note that the sampling based approach significantly improves performance.
